\documentclass[journal]{IEEEtran}

\ifCLASSINFOpdf

\else

\fi

\usepackage{moreverb,url}

\usepackage[colorlinks,bookmarksopen,bookmarksnumbered,citecolor=red,urlcolor=red]{hyperref}
\usepackage{graphicx}
\usepackage{amsmath}
\usepackage{latexsym}
\usepackage{amssymb}
\usepackage{fancyhdr}
\usepackage{array}
\usepackage{dirtytalk}
\usepackage{float}

\newcommand\labelname[1]{\say{\textit{#1}}}

\newcommand{\celine}[1]{\textcolor{black}{#1}}
\newcommand{\tim}[1]{\textcolor{black}{#1}}

\begin{document}

\title{Exploring to learn visual saliency: \\The RL-IAC approach}


\author{C\'eline~Craye\thanks{C\'eline Craye is a PhD research engineer at Thales SIX- Vision and Sensing laboratory. Email: celine.craye@ensta-paristech.fr},Timoth\'ee~Lesort\thanks{Timoth\'ee~Lesort is a PhD student at Thales SIX and at ENSTA Paristech, and a member of the INRIA FLOWERS team. Email: timothee.lesort@ensta-paristech.fr},~David~Filliat\thanks{David Filliat is a professor at ENSTA Paristech, and a member of the INRIA FLOWERS team. Email: david.filliat@ensta-paristech.fr},~and~Jean-Fran\c cois~Goudou\thanks{Jean-Fra\c cois~Goudou is a R\&I project manager at Thales SIX - Vision and Sensing laboratory. Email: jf.goudou@thalesgroup.com}\thanks{U2IS, ENSTA ParisTech, Inria FLOWERS team, Universit\'e Paris-Saclay, 828 bd des Mar\'echaux, 91762 Palaiseau cedex France}\thanks{Thales - SIX - Theresis - VisionLab 1, avenue Augustin Fresnel, 91767 Palaiseau, France}\thanks{You can visit the project's repository at \url{https://github.com/cececr/RL-IAC}}}% <-this % stops a space
\maketitle

\begin{abstract}

The problem of object localization and recognition on autonomous mobile robots is still an active topic. In this context, we tackle the problem of learning a model of visual saliency directly on a robot. This model, learned and improved on-the-fly during the robot's exploration provides an efficient tool for localizing relevant objects within their environment. 
The proposed approach includes two intertwined components. On the one hand, we describe a method for learning and incrementally updating a model of visual saliency from a depth-based object detector. This model of saliency can also be exploited to produce bounding box proposals around objects of interest. On the other hand, we investigate an autonomous exploration technique to efficiently learn such a saliency model. The proposed exploration, called \textit{Reinforcement Learning-Intelligent Adaptive Curiosity} (RL-IAC) is able to drive the robot's exploration so that samples selected by the robot are likely to improve the current model of saliency.
We then demonstrate that such a saliency model learned directly on a robot outperforms several state-of-the-art saliency techniques, and that RL-IAC can drastically decrease the required time for learning a reliable saliency model.
\end{abstract}


\section{Introduction}

In the scope of assistive robotics, where autonomous mobile robots assist and help humans with their everyday life tasks, the need for robots to efficiently analyze and understand their environment is critical. To this end, robots should have the capacity to efficiently find and identify objects they can interact with.

Object localization in cluttered environments is still a difficult problem. Today, deep learning-based methods provide efficient ways to localize and identify a large set of objects in a wide variety of complex configurations~\cite{krizhevsky2012imagenet}, but they generally require hours or days of offline training, high GPU resources, thousand to millions of training images, and are not really flexible to novelty. Furthermore, domestic mobile robots are meant to evolve essentially in indoor environments, interact with a limited amount of objects, for specific tasks and thus do not require such wide scope capacity. However, they should be able to adapt to novelty by quickly updating the representation of their environment. Learning to localize objects online and directly within the environment is then a very desirable property.

\label{sec:SaliencyLearning}
\begin{figure}
   \centering
      \includegraphics[width=.45\textwidth]{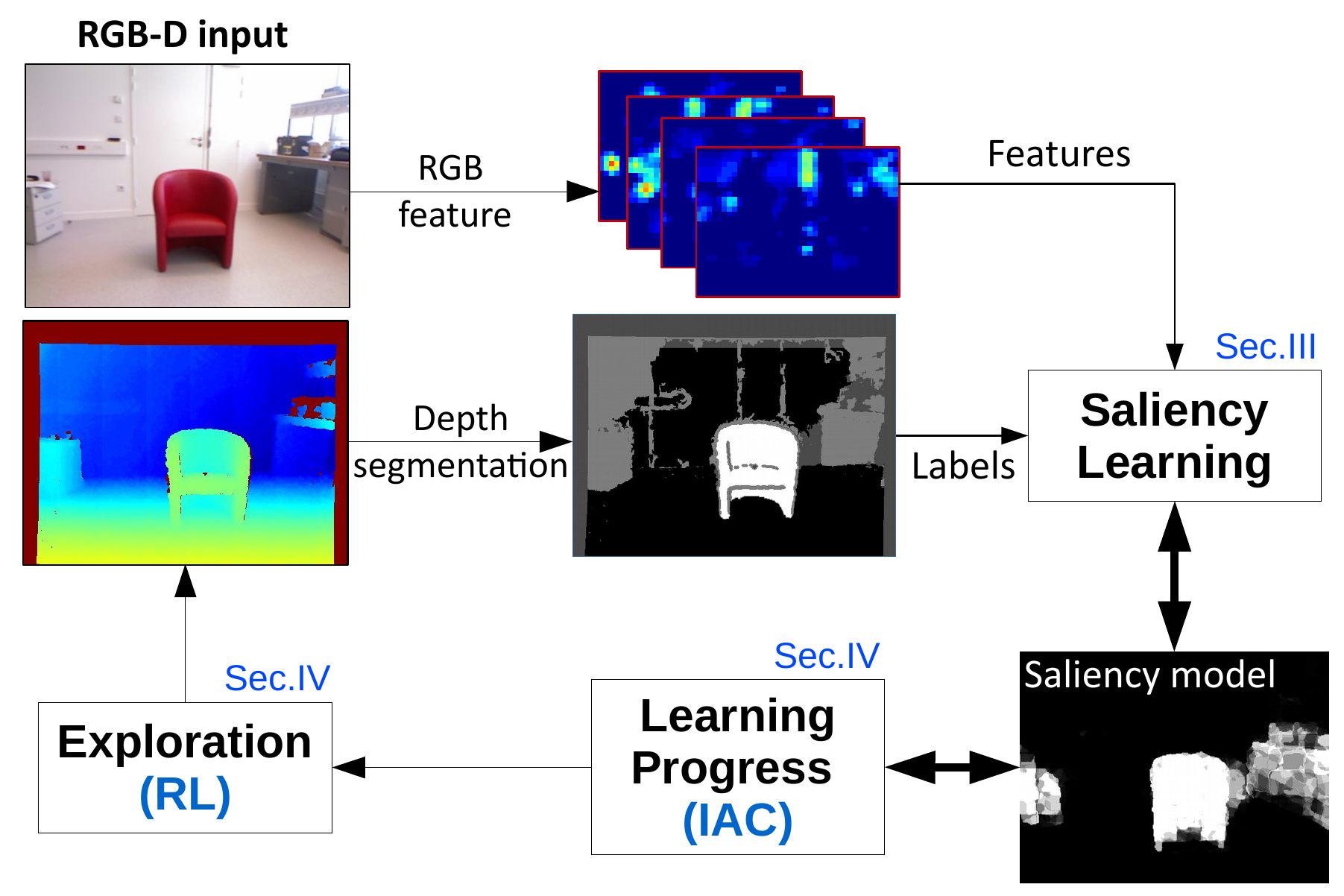}
      \caption{\textcolor{blue} Overview of our saliency learning and exploration approach.}
      \label{fig:generalArchitecture}
\end{figure}

Nevertheless, online learning must come with a methodical exploration of the environment. The displacement of the robot makes it possible to move to favorable observation conditions in order to improve recognition performances, but a critical point is to monitor this performance quality, and use this information to drive the robot accordingly.

In this article, we consider a mobile robot exploring its environment while building a model of visual saliency enhancing objects of interest. Based upon our previous work~\cite{craye2016RL-IAC}, we present a system (Fig. \ref{fig:generalArchitecture}) able to:

\begin{itemize}
\item produce object-oriented visual saliency maps from a model;
\item learn the saliency model incrementally directly within the robot's environment;
\item make the robot explore the environment autonomously and efficiently, by visiting in priority areas able to improve the saliency model.
\end{itemize}

More precisely, the system is composed of two major components: on the one hand, a method for learning a visual saliency model incrementally and without any user supervision. This model can be exploited to enhance objects of interest in the environment, and used as a posterior to generate bounding boxes around salient objects. On the other hand, we describe RL-IAC (\textit{Reinforcement Learning Intelligent Adaptive Curiosity}), an algorithm based on the \textit{Intelligent Adaptive Curiosity} that drives the robot in its environment, so that learning is done is an efficient and organized manner. RL-IAC encapsulates the saliency learning technique and can be seen as a whole system for autonomous exploration and efficient learning. We demonstrate that our method for learning saliency online generates saliency maps that are more accurate than most state-of-the-art technique in the robot's environment. In addition, the efficiency of RL-IAC for exploration is evaluated versus alternative environment exploration techniques. 

This article extends Craye \textit{et al.}~\cite{craye2016RL-IAC} by providing more technical details and experimental results. It also presents a new type of feature extractor based on convolutional neural networks, and a method to automatically and incrementally obtain a navigation graph required by the RL-IAC algorithm.\celine{The CNN feature extraction approach was already presented in a previous work \cite{craye18}, but the evaluation was carried out on a different dataset, and the use of RL-IAC was not investigated.}

The article is organized as follows: we present related work in Section~\ref{sec:RelatedWork}, Section~\ref{sec:SaliencyLearning} describes the method used to learn visual saliency incrementally, while Section~\ref{sec:RL-IAC} explains the exploration strategy based on RL-IAC. We propose an experimental evaluation of our system in Section~\ref{sec:ExperimentalResults}, and finally provide concluding remarks and perspectives in Section~\ref{sec:Conclusion}.

\section{Related work \label{sec:RelatedWork}}

As our system is based on two independent components, we consider separately the related work on saliency maps and object localization, and the one on the field of exploration on mobile robots. We last highlight our main contributions and positions towards state-of-the-art.

\subsection{Saliency maps and object localization}

To efficiently analyze visual inputs and interact with objects in cluttered environments, robots usually rely on a visual attention strategy. This mechanism turns the raw visual scene into selected and relevant information the robot should focus on, possibly involving zooming~\cite{minut2001reinforcement}, foveal vision~\cite{bjorkman2010active} or physical displacements~\cite{borji2010online,kragic2009object} of the robot. This concept has been widely studied and discussed~\cite{borji2013state,itti2001computational,frintrop2010computational}, from biological and computer vision points of view. We restrict visual attention in this study to the localization of objects of interest. 

Visual attention is strongly related with the concept of visual saliency, defined as a \say{subjective perceptual quality which makes some items in the world stand out from their neighbors and immediately grab our attention}~\cite{itti2001computational}. The first computational models of visual attention were relying on saliency maps~\cite{itti1998model}, representing the saliency of an image on a pixel-by-pixel basis. General convention is to associate a pixel intensity proportional to the pixel saliency. 

Saliency maps can be either purely \textit{bottom-up}~\cite{zhang2013saliency,erdem2013visual,hou2012image}, or refined by \textit{top-down modulation}~\cite{hamker2005emergence,zhao2011learning, frintrop2006vocus,frintrop2015traditional}. Bottom-up saliency highlights stimuli that are intrinsically salient in their context, which may sometimes be sufficient for scene exploration~\cite{zhu2012unsupervised}. However, top-down modulation, which highlights elements that are relevant for a specific task, is more meaningful for the problem of object detection in indoor environments. Saliency maps are either fixation-based~\cite{itti1998model,erdem2013visual} or area-based~\cite{cheng2011global,frintrop2015traditional,zhang2013saliency}. Fixation-based approach is related with the probability of a human being to make a fixation at a given pixel position, while area-based approach consider salient elements (typically objects) as a whole area of the image. The latter approach is then closely related to object segmentation. In the context of a mobile robot in an indoor environment, our technique aims to build top-down, object-oriented models of saliency.

Saliency maps are most of the time based on RGB images only, but a few of them also integrate the depth component~\cite{peng2014rgbd,ciptadi2013depth}. Another possible approach is to fuse both depth and RGB components~\cite{garcia2015saliency} processed separately. Usually, depth is good at detecting objects and is particularly well-suited for indoor environments. These approaches typically use geometrical constrains such as symmetries~\cite{potapova2014attention,ecins2016cluttered}, convexity~\cite{papon2013voxel}, or detecting elements placed on planar surfaces (such as floors or tabletops)~\cite{caron2014neural,ali2014contextual}. These approaches can detect objects much more accurately than using only the RGB component, but are limited by the sensor quality and geometrical constraints (reflectance, size or distance to the objects). Our approach uses the depth component to detect objects, and learns the visual aspect of these objects on the RGB image.

Machine learning, and especially deep learning have also been used for the generation of saliency maps. The best performance reported on saliency benchmarks is so far based on Convolutional Neural Networks (CNN).\tim{ Whether on stimuli-based benchmarks~\cite{2016arXiv161104849H, 2016ITIP...25.5012L} or object-based one~\cite{2016arXiv160301976L}, those methods have shown excellent results compared to traditional approaches. More interestingly, it was shown that CNN activation maps can be used as powerful objects detectors even when trained on a weakly-supervised basis~\cite{zhou2015cnnlocalization, oquab2015object}.
To obtain saliency based on deep learning, activation maps can be sampled from different CNN layers to produce multi-scale saliency maps~\cite{2016arXiv161104849H, 2016ITIP...25.5012L}. 
Moreover, these neural networks are also particularly efficient for image segmentation. Therefore, a CNN segmentation model can be used as a complement to a object detection CNN to improve the quality of the saliency maps \cite{2016arXiv160301976L}.
CNN models are generally learned by supervision but they can also be trained for image saliency prediction through adversarial examples~\cite{Pan_2017_SalGAN}. }

%SalGAN is an Adaptation of GAN architecture. It is trained with adversarial examples to predict an image saliency map with a deep CNN.
%~\cite{Pan_2017_SalGAN}

% DSS architecture is based on a CNN, introducing short connections in a skip-layer structure. It provides a multi-scale saliency maps, a property motivated by its necessity to perform segment detection.
%~\cite{2016arXiv161104849H}

%Visual Saliency Detection Based on Multiscale Deep CNN Features \cite{2016ITIP...25.5012L}
%Deep Contrast Learning for Salient Object Detection \cite{2016arXiv160301976L}

In the scope of object detection, the use of agnostic bounding boxes~\cite{alexe2012measuring,zitnick2014edge} has become a very popular alternative to the traditional sliding window technique. Not only this type of bounding boxes significantly decreases the number of windows to evaluate, but it also provides much more accurate bounding boxes. Several CNN-based object-detection techniques, such as faster R-CNN~\cite{ren2015faster}, are based on agnostic bounding box proposals. Hosang \textit{et al.} \cite{hosang2016makes} have published in that regard an extensive review on detection proposals. We investigate in this work how our saliency model can be used as a quality measure for these kind of proposals. 
\tim{Other methods take advantage of CNN to propose objects segmentation based on multi-scale features map ~\cite{Pinheiro15,Pinheiro16}.}

%DeepMask ~\cite{Pinheiro15} : propose a segmentation at several layers and make a mean of them to propose the final one.
%SharpMask ~\cite{Pinheiro16} : somehow learn to combine segmentations from different layers to produce the final one (same model as Deepmask for forward pass). "each pixel prediction is based on a complete view of the object". "We introduce a ‘refinement module 'R' that is responsible for inverting the effect of pooling and doubling the resolution of the input mask encoding" 

\subsection{Autonomous environment exploration}

In robotics, the \textit{exploration problem} is the one of maximizing knowledge over a working environment by means of a single or several robots. Autonomous exploration of the environment is then made possible by providing rules able to guide the robot's actions to reach specific goals in that regard. We here present three types of approaches that can be potentially combined to lead the robot's exploration.

\subsubsection{Vision-based exploration:} 
Exploration may be guided by visual inputs and driven by a \textit{visual attention mechanism}~\cite{itti1998model}. In this case a visual focus of interest is selected in the environment (from saliency maps computation, for example), and the actions performed by the robot aim to provide more information about the selected target. When the robot is equipped with PTZ cameras (pan-tilt-zoom)~\cite{canas2008overt,kragic2009object, minut2001reinforcement} or with a combination of contextual (wide field of view) and foveal (sharp field of view) cameras~\cite{bjorkman2010active, vijayakumar2001overt}, the actions are typically saccades, so that the zoomed camera is oriented towards informative regions. If the robot can move across its environment, actions are displacement in order to get a closer or better point of view of areas of interest~\cite{massios1998best,kragic2009object,lauri2014stochastic,borji2010online}. In all cases, exploration is based on a pre-attentive stage, where potentially relevant targets are selected and uninformative areas are discarded, and an attentive stage, where more complex tasks (such as grasping~\cite{kragic2006strategies} or object recognition~\cite{rasolzadeh2010active}) are performed on the targets to obtain more information about them.

\subsubsection{Map-based exploration:}
When autonomous exploration is made by a mobile robot, a 2-D or 3-D map of the environment is commonly used. Those maps can either be pre-defined before exploration~\cite{kragic2009object}, or incrementally updated~\cite{bazeille2011incremental} as the robot discovers new portions of the environment. The exploration problem in mobile robotics is often related with a problem of maximizing map under constrains. For example, minimizing displacement time~\cite{osswald2016speeding} while visiting a certain number of areas by solving a traveling salesman problem, minimizing the number of views~\cite{jebari2012combined} with an art-gallery problem, or re-visiting previously observed areas based on potential uncertainty reduction~\cite{kragic2009object} or information-gain~\cite{Santos2016}. Unlike visual attention strategies, exploration based on map coverage is often composed of a problem of displacement cost minimization.

\subsubsection{Skill-based exploration:}
When the robot aims to learn skills from its environment, actions can be oriented to the task of learning rather than that of pure exploration. This is then typically the case in reinforcement learning~\cite{barto1998reinforcement} (RL), where actions are taken to learn an optimal state-action policy rather than for gaining knowledge about the environment. \textit{Q-learning} is a typical RL algorithm for this kind of exploration~\cite{lopes2012exploration,minut2001reinforcement}. In the scope of developmental robotics~\cite{weng2001autonomous}, intrinsic motivations are also used as a drive for robot's acquisition of skills through experience and exploration. \textit{Intrinsic motivation}, defined as intrinsic reward (\textit{i.e.} not related to an external goal, but to the acquisition of competences or knowledge) able to drive a behavior, is a possible approach for guiding exploration in that regard. For example, Huang \textit{et al.}~\cite{huang2002novelty} have used \textit{novelty} to guide visual exploration, while Chentanez \textit{et al.}~\cite{chentanez2004intrinsically} have used the error of prediction of \textit{salient events} to speed up a classical reinforcement learning approach. To overcome limitations related to novelty or error in unlearnable situations, intrinsic motivation based on \textit{progress} has been proposed~\cite{nguyen2013learning,baranes2009r,schmidhuber2010formal}. The \textit{Intelligent Adaptive Curiosity} (or IAC)~\cite{oudeyer2007intrinsic} is one of the most emblematic implementation of intrinsically motivated exploration using progress. Learning progress has also been exploited in a reinforcement-learning context, typically with artificial curiosity~\cite{kompella2012autonomous,schmidhuber1991curious}, or to make exploration flexible to changes in the environment or wrong assumptions~\cite{lopes2012exploration}.

\subsection{Contributions}

So far, saliency maps are mostly used as black boxes and are not learned (although sometimes refined) directly during the exploration of a particular environment. Our first contribution is a method that incrementally learns saliency as the robot observes the environment. The produced saliency maps are therefore dedicated to the environment that was explored, but remain flexible to novelty. The model that is learned here is a top-down type of saliency, dedicated for generic robotics tasks (i.e. a saliency that detects objects the robot can interact with). The term saliency in this article is then more related to the concept of objectness, and the model that is learned is used to produce object-oriented saliency maps. We further show that these saliency maps can be used to refine object box proposals. Unlike most saliency techniques based on learning, ours is self-supervised, so that the robot is able to learn without any human annotation or assistance. The main mechanism consists in a transfer learning method between a restricted depth-based segmentation technique and the RGB frame.

Our second contribution is the RL-IAC (\textit{Reinforcement Learning Intelligent Adaptive Curiosity}) algorithm that provides an autonomous exploration strategy. RL-IAC encapsulates the online saliency learning technique, and uses it as a core component to drive the robot's exploration. RL-IAC is a hybrid approach: on the one hand, it uses the idea of the IAC algorithm~\cite{oudeyer2007intrinsic} that drives exploration towards learning progress. On the other hand, it uses a map-based exploration that tries to minimize the time spent in displacements. For that, we add a reinforcement-learning module to the traditional IAC approach. The RL module is constantly retrained during the exploration, so that the robot is attracted by regions with high learning progress while trying to minimize its displacements.

\section{Saliency learning and object proposals}

\label{sec:SaliencyLearning}
\begin{figure*}
   \centering
      \includegraphics[height = 7.2cm]{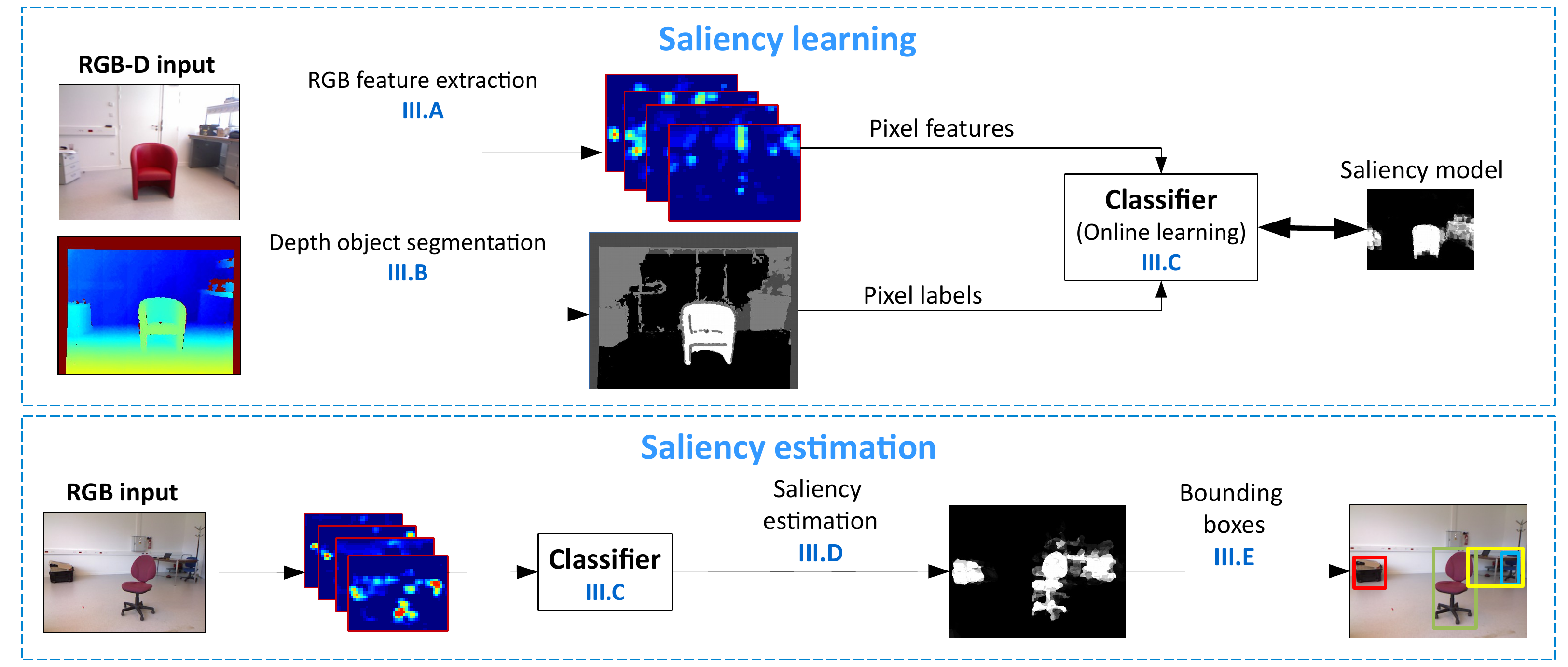}
      \caption{General mechanism of the saliency learning component. The RGB stream is used to generate CNN-based features, while the depth stream is sent to a segmentation algorithm. The segmentation result is used as a learning signal to train a classifier. The resulting trained model actually predicts the saliency of a given image, without using the depth segmentation component.}
      \label{fig:generalArchitectureSaliency}
\end{figure*}
 
This section describes the component that learns visual saliency and produces object proposals. Figure~\ref{fig:generalArchitectureSaliency} presents the general block architecture along with the corresponding section for each block. In a learning stage, the system extracts RGB features (see Section~\ref{sec:features}) and learns the visual (RGB) aspect of salient elements within their context using a depth-based object segmentation as a supervision signal (see Section~\ref{sec:segmentation}) that is only used in the learning phase. Learning is performed by a classifier (Section~\ref{sec:classification}) that produces and constantly updates a saliency model. When available, the saliency model is exploited to generate environment specific saliency maps using the RGB image only (Section~\ref{sec:saliencyEstimation}), and these saliency maps can be used to generate boxes that isolate objects of interest (Section~\ref{sec:bboxes}).

\subsection{RGB Feature extraction \label{sec:features}}

\celine{In this work, feature extraction is based on convolutional neural networks and is fully independant from the classification step. Considering deep learning frameworks, an end-to-end neural network architecture may be used for saliency learning, starting with convolutional layers
and ending up with fully connected one. However, metaparameters (learning rate, minibatch size, etc.) are hard to configure to allow an efficient incremental learning. A bad configuration could significantly deteriorate weights that were correctly learned for another problem. We then consider another approach to exploit deep neural networks: the use of the first layers of a well-trained network as a feature extractor. This way, we avoid instability problems, and the module is easily plugged in our architecture.}

We base our feature extraction upon the ideas of Zhou \textit{et al.}~\cite{zhou2015cnnlocalization}. In their article, a GoogLeNet architecture is used and fine tuned to perform object localization. The end of the network is replaced by a global average pooling layer, followed by fully connected layers providing strong localization capacities and trained on a weakly supervised dataset. This type or architecture is then able to produce, in some sense, class specific saliency maps. In addition, the weights of this network are publicly available.

We then use this available trained model and do not consider the layers after the global average pooling one. The feature extraction is done at the level of the \textit{class activation mapping}, or CAM layer (called \textit{CAM-CONV} in the network). This corresponds to the last fully convolutional layer of the network. According to Zhou \textit{et al.}, this layer is the one at which highly discriminative areas are enhanced.

\begin{figure}
   \centering
       \includegraphics[height = 6cm]{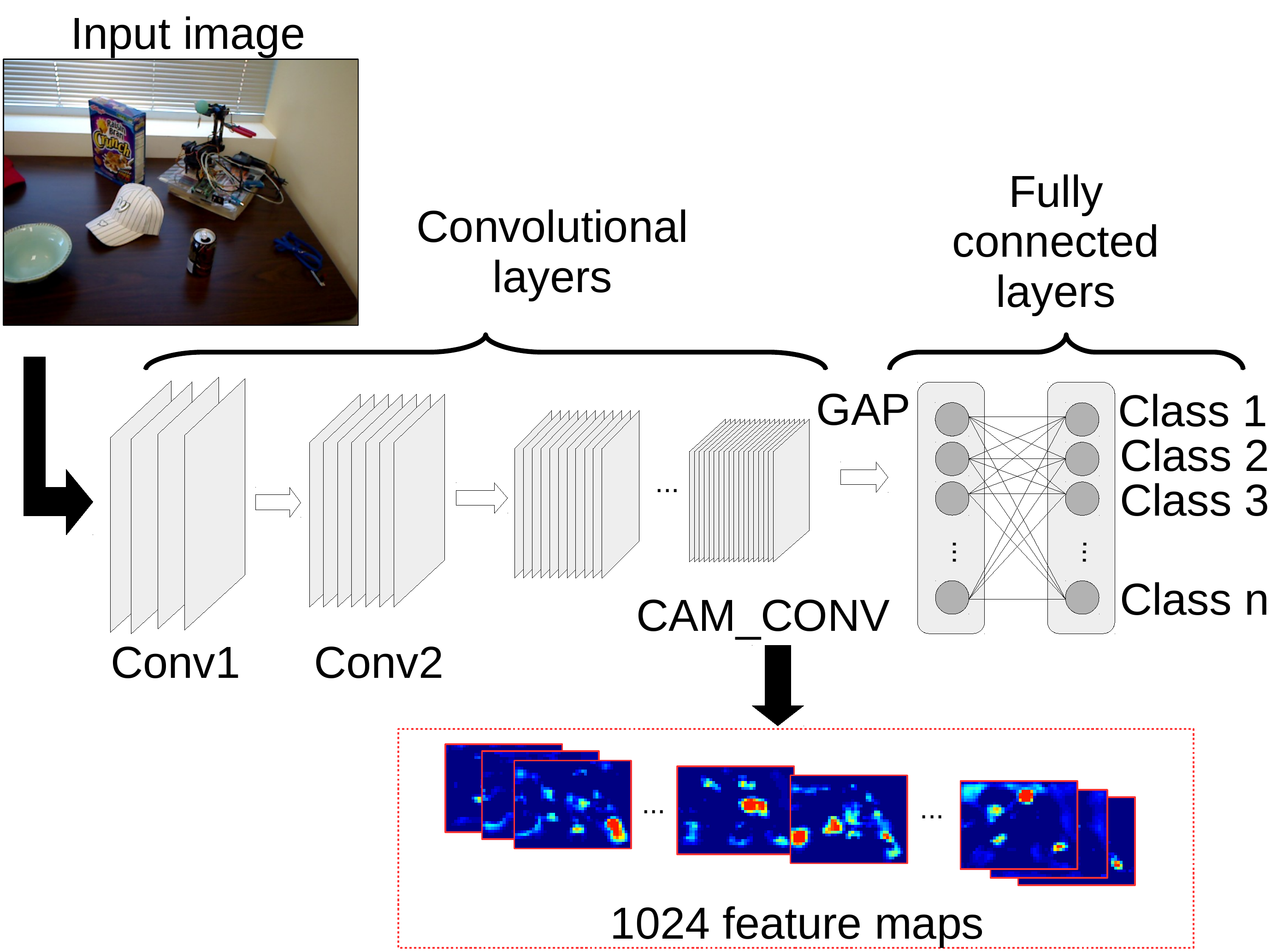}
      \caption{Example of feature maps extraction from an input RGB image (corresponding to section \ref{sec:features} in Figure~\ref{fig:generalArchitectureSaliency}).}
      \label{fig:featureExtraction}
\end{figure}

To get a set of feature maps, we then feed the network with the original RGB input image, without resizing it, and extract the 1024 maps of the CAM layer (See Figure~\ref{fig:featureExtraction}).
 
Because of striding and pooling in the network, the output feature maps have a resolution that is 16 times lower than the input image. To overcome this loss of resolution, we present in Section~\ref{sec:saliencyEstimation} a method to reconstruct saliency maps at the original scale.

\subsection{Depth-based object segmentation \label{sec:segmentation}}

\begin{figure}
   \centering
       \includegraphics[height = 7cm]{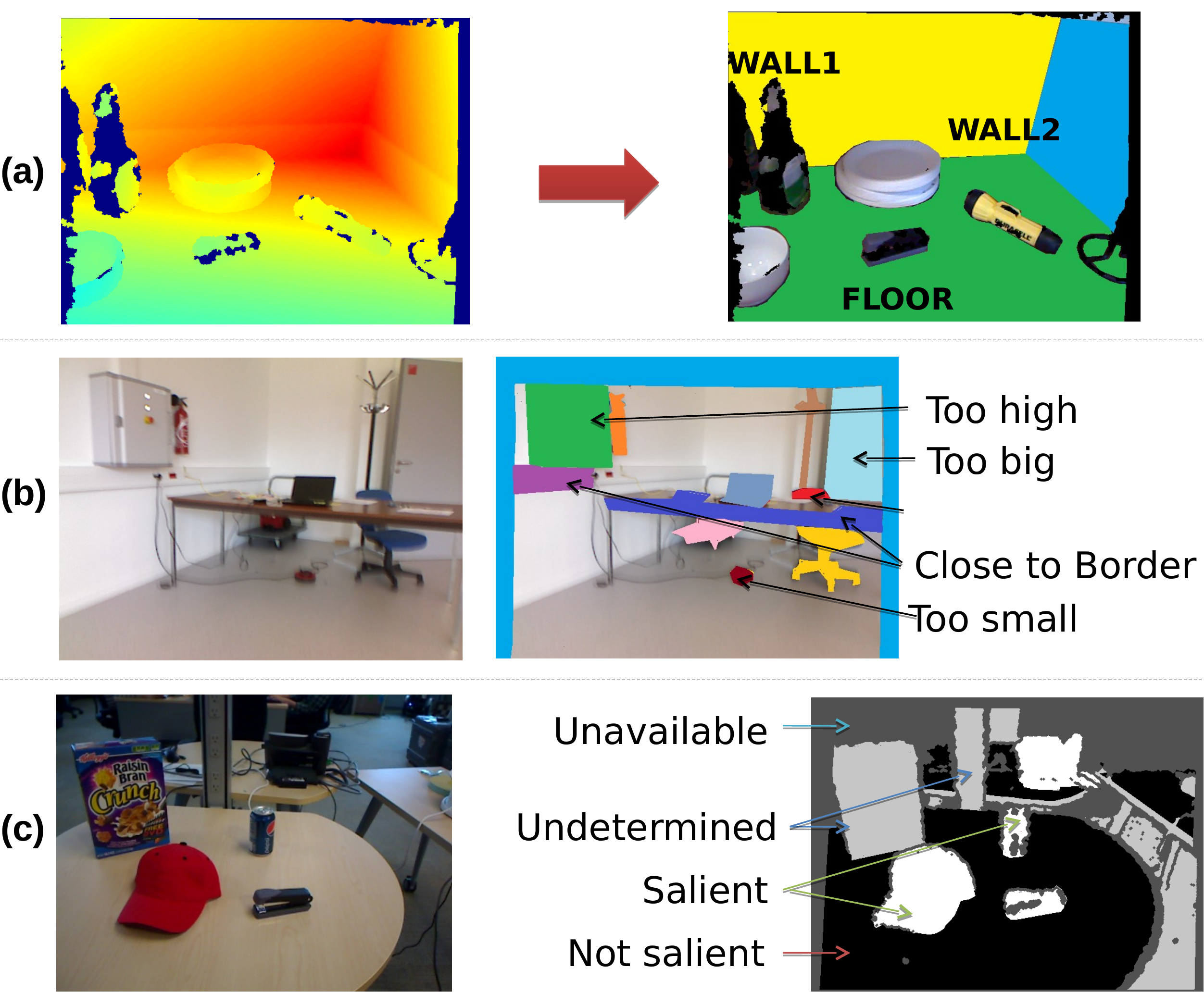}
      \caption{Main segmentation components (corresponding to section \ref{sec:segmentation} in Figure~\ref{fig:generalArchitectureSaliency}). (a) Floor and wall removal. (b) Object candidates filtering based on geometrical criteria. (c) Construction of the segmentation mask}
      \label{fig:segmentation}
\end{figure}

The segmentation procedure has been designed for RGB-D cameras such as Kinect or Asus Xtion, and mainly works for indoor environments to detect objects lying on planar surfaces (typically tables or floor), with a diagonal size between 10 and 180 centimeters. The method is a modified version of the depth-based object segmentation of Caron \textit{et al.}~\cite{caron2014neural} and is used to produce a partial but accurate estimation of the salient objects in the scene. Processing is based on the depth-map only. Figure~\ref{fig:segmentation} illustrates the main components of this approach.

As a first step, the depth map is turned into a point cloud, and the algorithm detects the major plane (most likely the floor, or a tabletop) of the cloud based on a RANSAC algorithm~\cite{fischler1981random}. The major plane is then tracked in the following frames and during the whole sequence to make sure that the same surface is used during the whole experiment and to monitor false detections.

Given this major plane, potential walls are detected and filtered out: they are detected by finding large planar surfaces perpendicular to the major plane. The remaining points of the cloud are likely to be part of salient objects, but could as well be small portions of walls, poles, or any other artifact that is irrelevant for a robot. Remaining points are grouped in blobs to form object candidates. We then remove candidates that are either too small, too large, or too far from the floor. Last, to avoid false detection from large objects cut by the border of the frame, only candidates having no contact with the border of the field of view are kept and considered as salient objects. Whether discarded or not, all of the object candidates are associated with a bounding box to generate the SegBoxes (see Section~\ref{sec:bboxes}).

To convert this segmentation procedure into a 2-D segmentation mask compatible with saliency learning, we project the whole point cloud back to the image frame and attribute a label value to each corresponding pixel. Figure~\ref{fig:segmentation}, row (c) illustrates an example of such segmentation mask. In this case, pixels having no corresponding point cloud, because of reflectance (for example, the plastic bottle in (a)), shadows (at the border of objects), or visible points too far from the kinect are labeled \labelname{unavailable} (dark gray). Points of the cloud that either belong to the major plane or to a wall are labeled \labelname{not salient} (black). Points of the cloud belonging to a cluster detected as a salient object is labeled \labelname{salient} (white). Last, all remaining points are categorized as \labelname{undetermined} (light gray), as the algorithm was not able to determine their actual state of saliency. In particular, candidates that are too close to a border could be either walls or objects (see portion of wall and table of row (b) that are both close to a border of the image). 

\subsection{Online learning}
\label{sec:classification}
\begin{figure}
   \centering
       \includegraphics[height = 4.1cm]{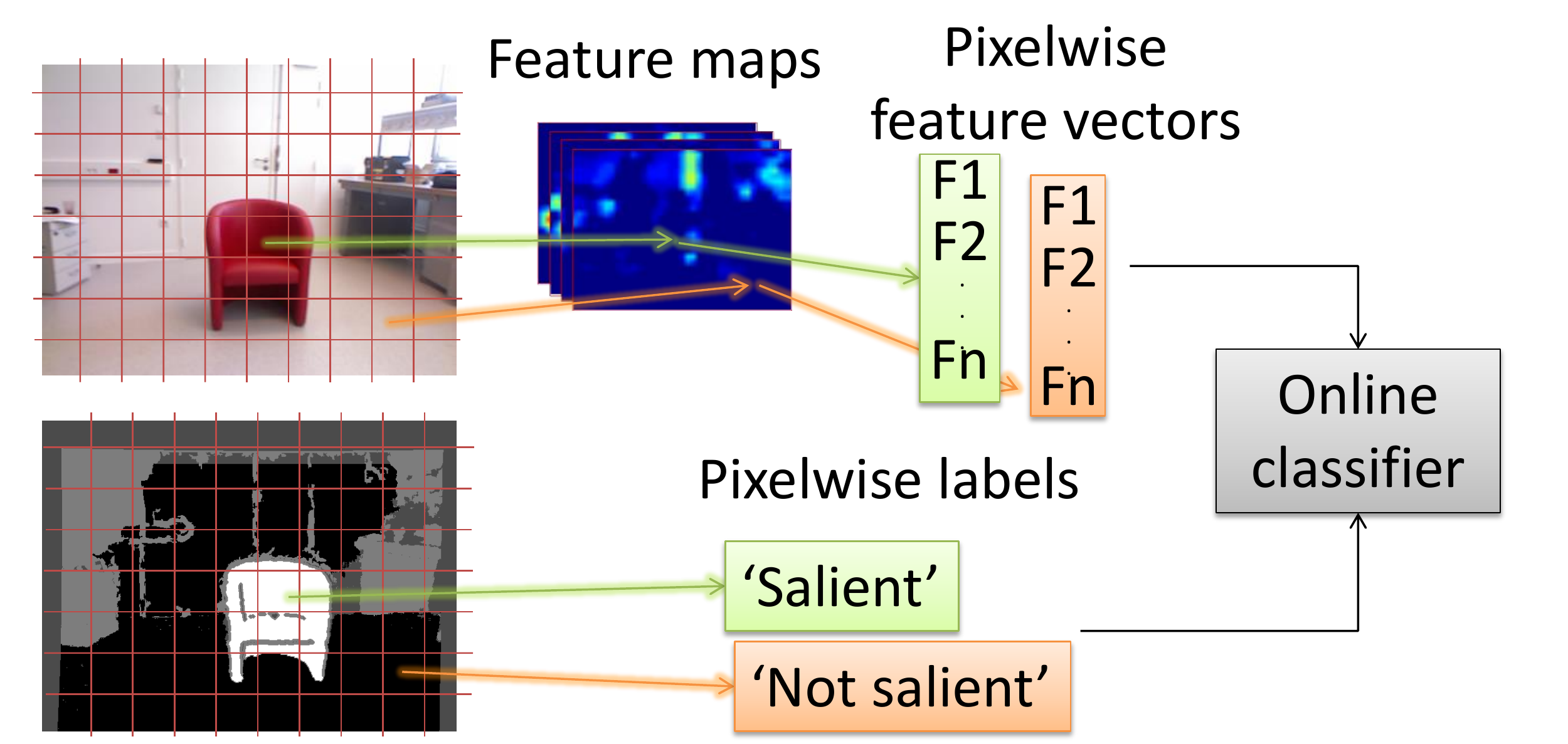}
      \caption{Pixelwise data formatting for online saliency learning.}
      \label{fig:dataFormatting}
\end{figure}

Learning is made possible by the feature maps, the segmentation mask as a learning signal, and an online classifier. The classifier is continuously updated based on the RGB-D observations, turned into a set of labels and features: the segmentation mask is first resized to the same size as the feature maps. The 1024 features associated to each pixel are collected and turned into a feature vector so that each pixel is attributed a features-label sample (see Figure~\ref{fig:dataFormatting}). These samples constitute the dataset send to our classifier to train our saliency model. To make sure that the dataset has as few false detections as possible, only pixels having a label \labelname{salient} or \labelname{not salient} in the segmentation mask are selected to feed the saliency model.

The classifier used in our implementation is a random forest, which is not originally designed for online training. No available version \cite{saffari2009line,lakshminarayanan2014mondrian} of online random forest was satisfying in terms of speed and performance, so that we adapted the offline version to make re-training fast enough. To this end, a limited number of samples from each new frame is used to update the classifier: for each new frame, we randomly pick up 500 pixels and add the corresponding samples to a dataset cumulated from the beginning of the experiment. Then, we update the classifier by only re-training a small fraction of the forest at a time: we randomly select 4 trees among the 50 in the forest, and we re-train them with 70\% of the whole dataset. To limit the size of the dataset, we fix a maximum number of 100 000 samples. If the dataset exceeds this size, we randomly remove samples to meet the maximum size requirement. As a result, after each update, the classifier is able to estimate the saliency of an input based on the model trained with the previous observations, and the RGB image only.

\subsection{Saliency estimation}
\label{sec:saliencyEstimation}
Saliency maps are generated by applying the classifier to RGB images. To this end, features are extracted from an input RGB image and are sent to the classifier to produce a saliency evaluation. For each pixel of the feature map, the classifier outputs a probability of the pixel to be salient. The output score is then a value between 0 (\labelname{not salient}) and 1(\labelname{salient}) corresponding to a fuzzy state of saliency (see for example Figure~\ref{fig:boundingBoxes} where saliency maps are represented as heatmaps). To rescale this low-resolution saliency map to the original input image size (recall that the deep feature extraction downsamples the image by a factor of 16), we generate SEEDS superpixels~\cite{van2012seeds} from the original image (350 superpixels for 640$\times$480 images in our experiments). We associate a low-resolution pixel to each superpixel by finding the pixel that would be the closest to the superpixel centroid if rescaled at the same size. The saliency value estimated for this pixel is then used to cover the entire superpixel. 

Although less accurate than depth segmentation, saliency maps provide an estimation of the saliency for each pixel of the image, as opposed to the partial saliency prediction of the segmenter. The classifier is also able to generalize saliency even when the segmenter fails at predicting it.

\subsection{Object bounding box proposals}
\label{sec:bboxes}
\begin{figure}
   \centering
       \includegraphics[height = 4.7cm]{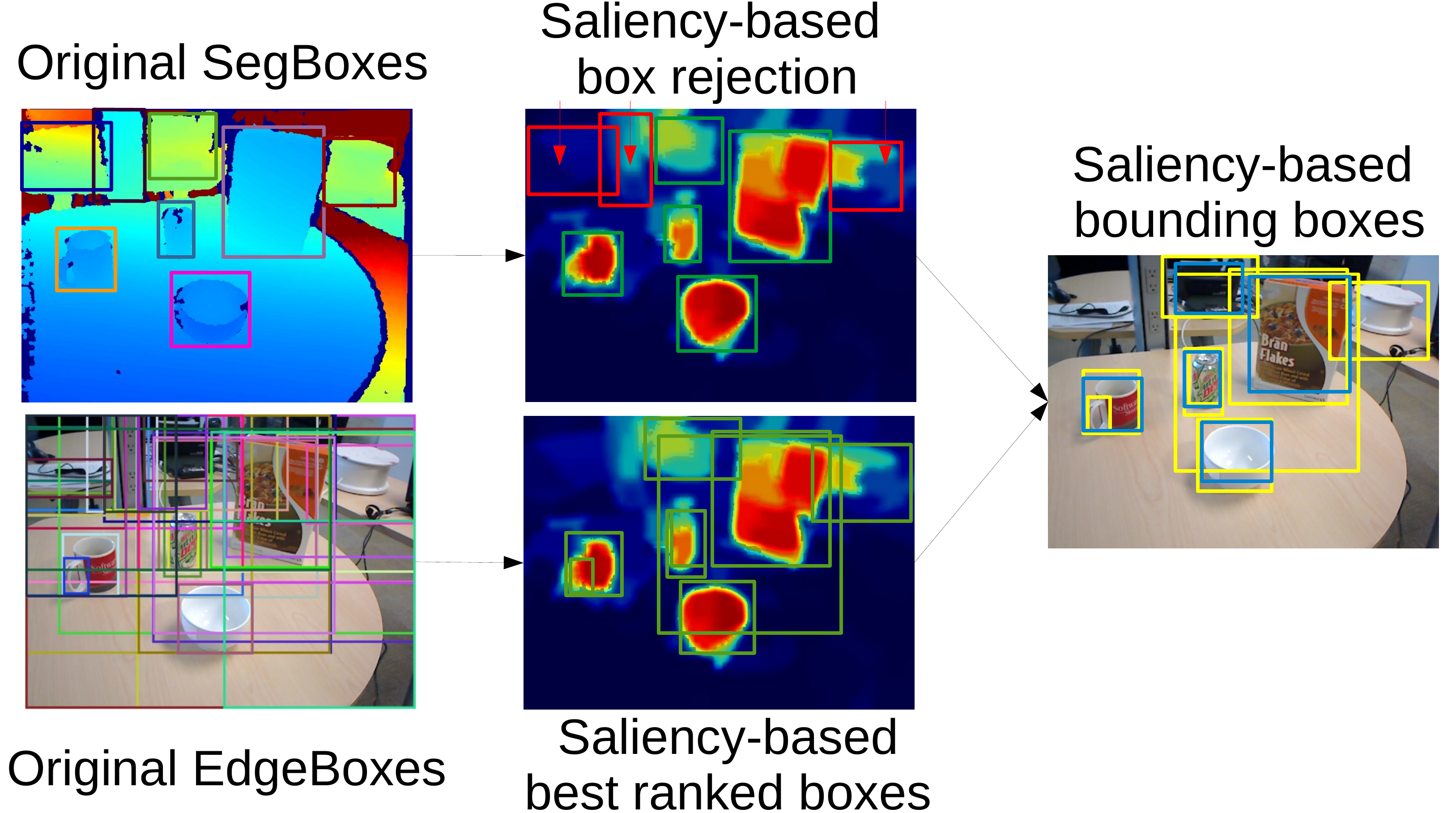}
      \caption{Bounding box proposals biased by our saliency map}
      \label{fig:boundingBoxes}
\end{figure}

The saliency map provides an indication of the interestingness of a given pixel, but does not say much about objects. In order to localize objects in an image, an additional step is then necessary to group salient pixels into object candidates, represented in Figure~\ref{fig:boundingBoxes}. To this end, we use two types of bounding box proposals, and we select or reject each of them based on a score related to saliency. The first bounding boxes are obtained by the EdgeBoxes \cite{zitnick2014edge} algorithm: we compute for a given RGB input the 100 most likely EdgeBoxes, and their associated edge score ($h_b^{in}$ in \cite{zitnick2014edge}). The second type of bounding boxes are obtained with the segmentation result (called SegBoxes here for simplicity): the segmentation algorithm produces, on top of salient objects, some additional object candidates.during segmentation, pixels of the depth map are clustered in order to create object candidates (See Section \ref{sec:segmentation}). Among these candidates, some are labeled \labelname{salient}, some should be salient but are labeled \labelname{undetermined}, and some are just artifacts. We then define Segboxes as the bounding boxes around all objects candidates proposed by the output segmentation mask. of the segmentation mask.

For both EdgeBoxes and SegBoxes, we associate each box $B$ with a score related to saliency (called here the \textit{saliency consistency score}, or SCscore), representing the ratio of salient pixels in the box: 
\begin{equation}
 SCscore(B) = \frac{1}{w_B\times h_B}\sum_{i,j \in B}S(i,j)
 \label{eq:sal_consistency}
\end{equation}
where $S(i,j)$ is the saliency of the pixel at $(i,j)$, obtained from the saliency map and $w_B$ and $h_B$ are the width and height of $B$. The highest the score is for a given box, the most likely it is to contain a salient object. For the EdgeBoxes, the SCscore is multiplied by $h_b^{in}$. This way, small boxes found within a salient object might be rejected if the $h_b^{in}$ score is low enough. Last we filter out Segboxes and Edgeboxes with a final score below a certain threshold and keep the remaining ones. In our dataset, we found $SCscore=0.2$ for SegBoxes and $SCscore\times h_b^{in}=0.01$ to be good trade-off thresholds between false alarm and false rejection rates.

\section{RL-IAC}
\label{sec:RL-IAC}

Our exploration strategy, that we call \textit{Reinforcement Learning-IAC} (or RL-IAC) is based on the IAC (\textit{Intelligent Adaptive Curiosity}) algorithm. In IAC, the robot focuses on areas where learning is neither trivial nor impossible. This way, learning is faster as no time is wasted in useless or unlearnable areas, and better as mainly relevant samples are selected. Nevertheless, the original version of this algorithm does not consider the case where actions (displacements in our case) have a non negligible time. To make it compatible with our application, we couple IAC with some navigation information to find a right balance between learning and displacement.

Further explanations about the differences and similarities of our approach with traditional IAC applications have been described in a previous work~\cite{craye2016use}. We focus here on describing the mechanisms of RL-IAC as a whole.

\begin{figure*}
   \centering
       \includegraphics[height = 7cm]{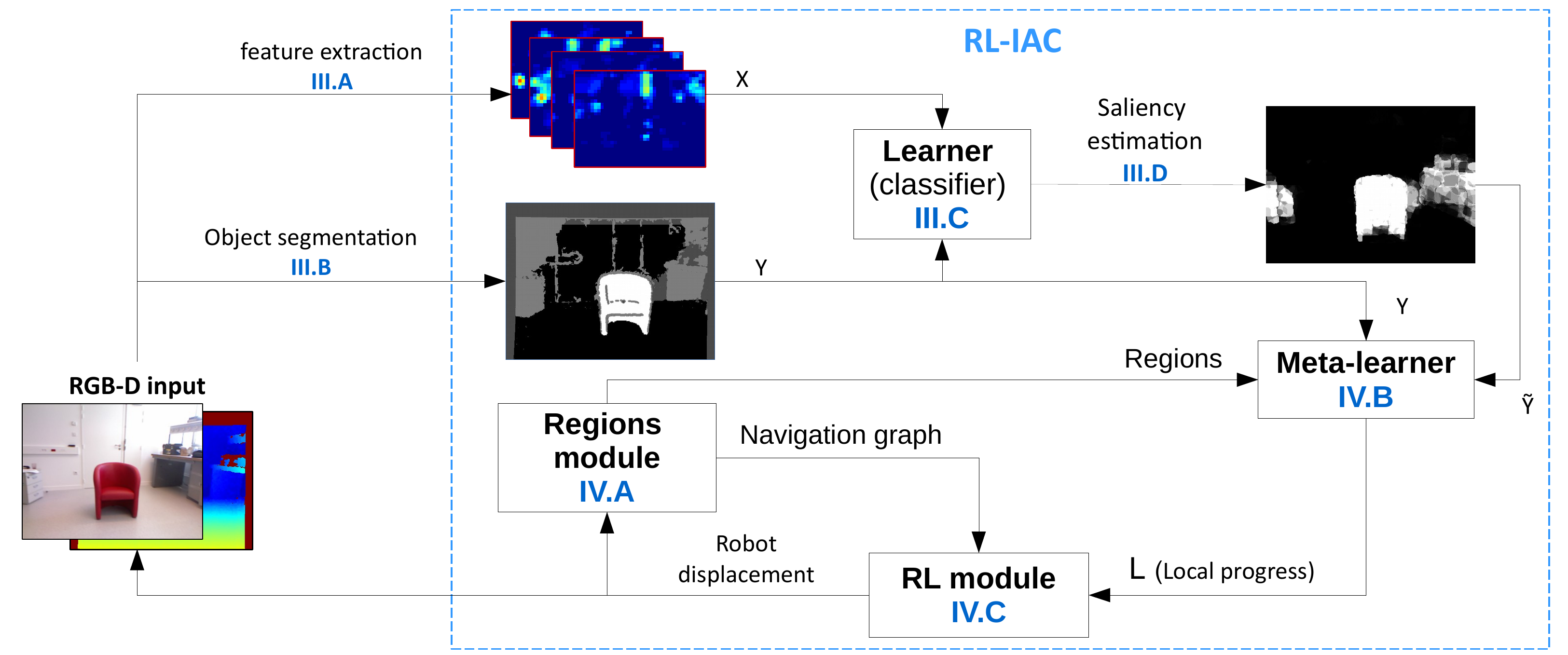}
      \caption{General achitecture of the RL-IAC module. The saliency learning module is improved by using a feedback loop considering the learning quality (through meta learning) in order to guide the robot's displacements.}
      \label{fig:generalArchitectureRLIAC}
\end{figure*}

Figure~\ref{fig:generalArchitectureRLIAC} presents the general architecture of RL-IAC, using the notations and vocabulary proposed by Oudeyer \textit{et al.}~\cite{oudeyer2007intrinsic}. The module strongly depends on the saliency learning module and takes as an input the feature maps $X$ and the object segmentation $Y$ to feed a learner (called classifier in Section~\ref{sec:classification}) that is constantly updated. Based on the available model, the learner produces a saliency map, which is also an estimate $\widetilde{Y}$ of segmentation $Y$. In the meanwhile, as the robot navigates in its environment, the space is cut into regions based on a 2-D map, and a navigation graph is updated to model the connexity between regions (Section~\ref{sec:regions}). By comparing $Y$ and $\widetilde{Y}$, the meta learner produces a local estimation of the error in each region, and derives a local progress measure $L$ (Section~\ref{sec:metaLearner}). Last, a reinforcement learning module (Section~\ref{sec:RL}) uses the navigation graph as the set of states and actions, and attributes to each state the reward $L$ of the corresponding region. The RL module is re-trained at each step to provide the robot with a displacement policy. After each displacement, a new RGB-D input is acquired and the modules are updated. \celine{The bounding box proposal module (section \ref{sec:bboxes}) was left out of the diagram as it was not exploited in the RL-IAC approach.}

\subsection{Regions and navigation graph}
\label{sec:regions}

\subsubsection{Role in RL-IAC}
One of the essential component of RL-IAC is to locally monitor how good the learner is at predicting saliency. This local estimation is obtained by creating statistics on samples collected within the same region. In our case, regions are defined as portions of a 2-D map of the environment, produced by a SLAM algorithm. The $(x,y)$ position of the mobile robot on the map then determines the region that is currently being explored. 

In addition, displacement in the environment is made possible by a navigation graph, representing the different regions, their relative positions and connexities, and the distance between two neighbor regions centroids. This navigation graph is also used by the reinforcement learning module to decide the next displacement of the robot.

\subsubsection{Algorithm:}
\begin{figure}
   \centering
       \includegraphics[height = 6cm]{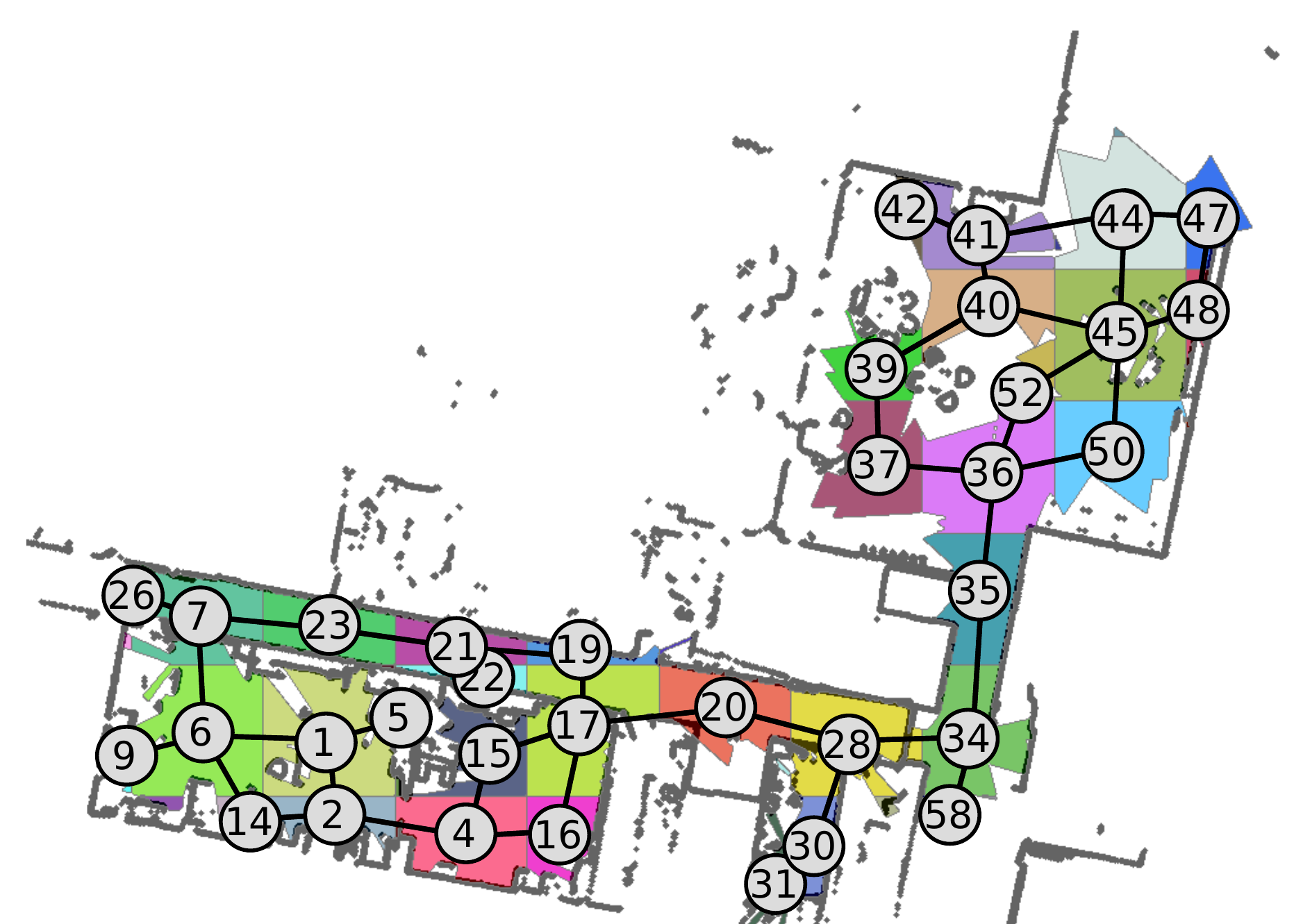}
      \caption{Regions and navigation graph obtained after the exploration of the ENSTA building. Note that the bounding box proposal is independent from the RL-IAC process, so that it is not displayed here.}
      \label{fig:navigationGraph}
\end{figure}

We propose a method that incrementally splits the space into regions and produces the associated navigation graph. The method relies on an occupancy grid (the 2-D map) of the environment and the visible field of view of the RGB-D sensor. Figure~\ref{fig:navigationGraph} presents an example of regions and navigation graph obtained this way. On this figure, visible areas are colored depending on the region they belong to. Walls and obstacles are represented by gray pixels. Lastly, circles with region index correspond to the region centroid, and edges represent the available displacements of the robot. Each region has at most 4 neighbor regions.

Initially, the occupancy grid has a pre-allocated size, where each pixel state is \labelname{unexplored}. This map is first divided in proto-regions based on a regular grid of arbitrary size of 5 meters-length. This way, each position of the occupancy grid is associated with a proto-region. The regions determined in our algorithm are subdivisions of those proto-regions. 

\begin{figure}
   \centering
       \includegraphics[width = 8cm]{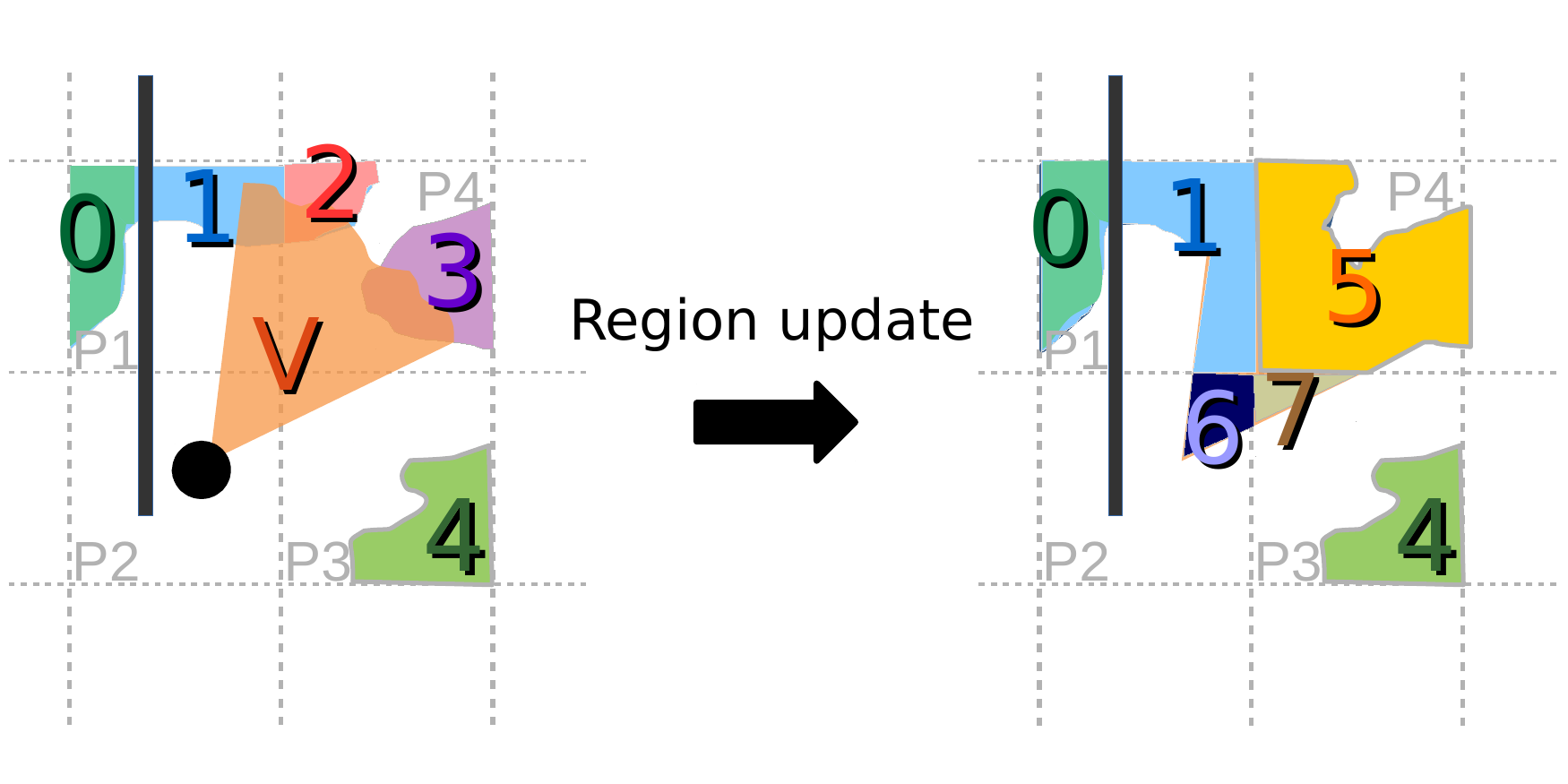}
      \caption{Update procedure. the robot is in proto-region $P_2$ but the field of view $V$ covers proto-regions $P_1$ to $P_4$. In $P_2$, no regions were previously defined, so a new region is created (region 6). In $P_3$, region 4 already existed but has no overlap with $V$: a new region is created (region 7). Region 1 in $P_1$ overlaps $V$, so the region is extended. Lastly, $V$ overlaps two regions in $P_4$, so that regions 2 and 3 are merged to create region 5. Last configuration, when a wall is splitting a proto-region, two separate regions are naturally created on each side (as they are not connected components). Thus, region 0 and 1 of $P_1$ are separated by a wall and cannot be merged together.}
      \label{fig:regionsUpdate}
\end{figure}

As soon as a new robot observation is available, we update the occupancy grid based on the RGB-D field of view: we transpose the point cloud obtained from the depth map in the occupancy grid frame. Points belonging to the floor are marked as \labelname{free} on the occupancy grid, and other visible points are marked as \labelname{occupied}. Let us denote $V$ the list of visible points marked as \labelname{free} on the map. Let us consider $ \{P_i\}_{i = 1 .. N} $ the $N$ proto-regions having an overlap with $V$. Each $P_i$ is then the list of all pixels contained in the square delimiting the proto-region. For a given $P_i$, let us define $ \{R_{j}\}_{j = 1 .. M} $ the pixels of the $M$ regions contained in $P_i$. Regions are now updated using the following procedure for each $P_i$ (Please refer to Figure~\ref{fig:regionsUpdate} for an illustration of each case):
\begin{itemize}
\item If $V\cap P_i\cap\{R_{j}\}_{j = 1 .. M} = \emptyset$, create a new region $R_{M+1}$ constituted with all pixels of $V\cap P_i$;
\item If $V\cap P_i \cap\{R_{j}\}_{j = 1 .. M} \neq \emptyset$ and $V$ is overlapping a single region $R_{j}$, update $R_{j}$ with $V$. $R_{j} \longleftarrow R_{j} \cup V\cap P_i$  
\item If $V\cap P_i \cap\{R_{j}\}_{j = 1 .. M} \neq \emptyset$ and $V$ is overlapping a set of $L \geq 2$ regions $\{R_{k}\}_{k \in 1 .. L}$ merge all those regions by creating a region $R_{M+1} = V\cap P_i\cup \{R_{k}\}_{k \in 1 .. M}$. Then, empty all $R_k$.
\end{itemize}

The nodes of the navigation graph correspond to the positions of the regions centroids (if not empty). Edges are determined based on the connexity between regions and their neighbours. In other words, if a region has a common border with another region, they are connected by an edge. Note that regions within the same proto-region do not have any connexity, otherwise they would have been merged. As a results, edges are necessarily between a region of proto-region $P_i$, and a region of the upper, lower, left or right proto-region $P_j$.
We also attribute to each edge a weight representing the distance between the two centroids.

\subsection{Meta-learner}
\label{sec:metaLearner}

The meta learner aims to monitor the local error made by the learner and derive a local estimate of the learning progress. The local estimation is made possible by grouping samples collected in the same regions and making statistics within each region. Recall that the robot is in region $R_i$ at time $t$ if its current position $(x(t),y(t))$ falls within $R_i$'s boundaries.

Error estimation is done by comparing a segmentation mask (the learning signal) with the corresponding saliency map (the estimated signal). More precisely, from an RGB-D input, we extract features and compute a segmentation mask. We then consider the observation set $O$ by keeping only \textit{salient} and \textit{not salient} pixels from the segmentation mask. We estimate the saliency response for each of these pixels. These responses (between 0 and 1) are then binarized with a threshold of 0.5 to obtain the estimation set $E$. Last, we compute the estimated learning error $Err$ of a particular frame based on Equation~\ref{eq:error_rate}:
\begin{equation}
Err = 1-F_1(O,E)
\label{eq:error_rate}
\end{equation}
where $F_1(.,.)$ is the $F_1$ score \footnote{ $F_1 = \frac{2tp}{2tp+fp+fn}$, where $tp$, $fp$ and $fn$ are the true positives, false positives and false negatives. We use the $F_1$ score as our error metrics, because \textit{not salient} pixels are representing more than 80\% of the samples, making accuracy inappropriate for error estimation.}.

The meta-learner stores a history of the learning errors for each regions. If the robot is in region $R_i$ and acquires an RGB-D input, the error associated to this observation is then added to the history of $R_i$. Supposing that this input is the $n$th observation in this region, we denote as $Err_i(n)$ the associated learning error.

The estimation of the learning progress in region $R_i$, is obtained by exploiting the error history sequence. For that, we apply a linear regression of the error history over the last $\tau$ samples ($\tau=10$ in our case):
\begin{equation}
\begin{pmatrix} Err_i(n-\tau) \\ \vdots \\ Err_i(n) \end{pmatrix} = \beta_i(n) \times \begin{pmatrix} n-\tau \\ \vdots \\ n \end{pmatrix} +  \begin{pmatrix} \epsilon(n-\tau) \\ \vdots \\ \epsilon(n) \end{pmatrix}
\label{eq:progress}
\end{equation}
with $\epsilon(n)$ the residual error and $\beta_i(n)$ the regression coefficient. $\beta_i(n)$ then represents the derivative of the learning error after $n$ observations. The learning progress being defined as the derivative of the learning curve (opposite of learning error), we obtain the  progress $LP_i(n)$ in region $R_i$ by Equation~\ref{eq:progress2}: 
\begin{equation}
LP_i(n) = \frac{2}{\pi} |atan(- \beta_i(n))|
\label{eq:progress2}
\end{equation}
We transform the slope $\beta_i$ with an arc tangent to have the learning progress normalized between -1 and 1, and we consider the absolute value of the arc tangent to force the robot to explore regions where learning is decreasing as well.

\subsection{RL-based displacement policy}
\label{sec:RL}

In previous applications of IAC, the time required to reach a region $A$ when in a region $B$ is not considered. In this scenario, a greedy policy exploring the regions with highest learning progress is enough. For a mobile robot moving in a large environment (e.g. a building), the displacements between two regions can be extremely time consuming, making the greedy policy inefficient. 

We therefore represent the regions by the navigation graph described in Section~\ref{sec:regions} that encodes the relative distances between regions and the possible displacements the robot can make to reach a region. We use this navigation graph to model states, actions and rewards of a reinforcement learning problem (we use Q-Learning \cite{watkins1992q}). The reward is the learning progress in each region, and a batch of simulations is run on this setup to determine the policy that optimizes progress while minimizing displacements. After simulations, the robot moves to another region by following this policy.

To describe how the module work, let us consider that the robot is in a region $R_i$ at time $t$. As the navigation graph and learning progress is constantly evolving, the procedure is repeated before each robot's displacement.
\subsubsection{States:}
The states are the node of the navigation graph. In other words, the regions centroids.
\subsubsection{Actions:}
Actions are represented by displacements on the graph. Each region is connected by edges with neighbor regions in one of the four (above, below, left and right) adjacent proto-regions. As a result, the robot selects one among actions \labelname{up}, \labelname{down}, \labelname{left} or \labelname{right}, depending on the graph edges, and moves to the  corresponding neighbor region.
\subsubsection{Reward:}
A reward $r$ is associated with each region $R_j$. We take for each region the last calculated learning progress. Thus, at time $t$, if $R_j$ contains $n_j(t)$ observations, we define reward as:
\begin{equation}
r(R_j) = LP_j(n_j(t))
\label{eq:reward}
\end{equation}
$LP_j$ being the learning progress defined by Equation~\ref{eq:progress2}.
\subsubsection{Simulation setup:}
We simulate a batch of 1000 episodes where a virtual robot moves in the navigation graph. For each episode, the initial state is $R_i$ (the actual state of the robot). During the episode, reward is collected right after taking an action and arriving in a given region. The episode stops when the traveled distance exceeds $N = 1km$ \footnote{This distance is obtained by cumulating the edge weights visited by the robot during the simulation.}.
\subsubsection{State-Action policy:}
During the batch of episodes, a state-action matrix is updated according to the following rule  
\begin{equation}
Q(s_k, a_k) = r(s_k) + \gamma \underset{a'\in A}{\mathrm{Max}}Q(s_{k+1}, a')
\label{eq:Qlearning}
\end{equation} 

with $\gamma$ the discount factor (0.9 in our implementation), $s_k$ the region where the robot is after $k$ (virtual) actions, $a_k$ the action to take next, $A$ the set of all possible actions, and $s_{k+1}$ the region after taking action $a_k$.
During the whole simulation, the virtual robot follows an $\epsilon$-greedy policy ($\epsilon=0.1$) to take the next action.
\subsubsection{Robot displacement:}
Once all the episodes have been simulated, we use the Q-matrix to select the next (not virtual) region to visit. For that, we select the next action $a_t$ that should be taken by the robot according to Equation~\ref{eq:bestQ}:
\begin{equation}
a_t = \underset{a'\in A}{\mathrm{Argmax}} (Q(R_i, a'))
\label{eq:bestQ}
\end{equation}
10\% of the time, the policy is not followed and the action is selected randomly among all available actions. 

We now consider the region $R_j$ connected to $R_i$ and accessible from action $a_t$. A position $(x_j,y_j)$ in $R_j$ is randomly selected and constitutes the next target to reach. The robot then moves to this position, updates the navigation graph, grabs a new RGB-D input, updates the learner and meta-learner, and determines by Q-learning the next position to reach.

Note that each Q-learning policy is obtained by considering the navigation graph and the reward as constant during the whole simulation. This assumption is not representative of the real world, as each displacement influences both the regions and the learning progress (that would eventually decrease to 0 when the learner cannot be any better). However, the assumption is accurate enough to suggest a displacement. As the Q matrix is re-estimated before each new action, this approximation does not introduce a significant bias. Moreover, to force the robot to quickly get a first estimation of the progress in each region, we force the progress in a given region to be very high as long as less than three samples are collected in that region. This additional constraint has the same effect as the R-MAX \cite{brafman2003r} exploration policy. 

\section{Experimental results}
\label{sec:ExperimentalResults}

We evaluate in this section the saliency learning and the RL-IAC exploration strategy separately, on datasets constructed slightly differently. Experiments were carried out on both publicly available datasets and datasets recorded on a mobile robot in our laboratory.

\subsection{Saliency maps and object proposal}
\label{sec:salLearningEval}

In this Section, we use three different datasets containing RGB-D images and ground truths, to evaluate both our saliency maps and the bounding box proposals methods.

\begin{figure*}
\centering
      \includegraphics[height = 8.2cm]{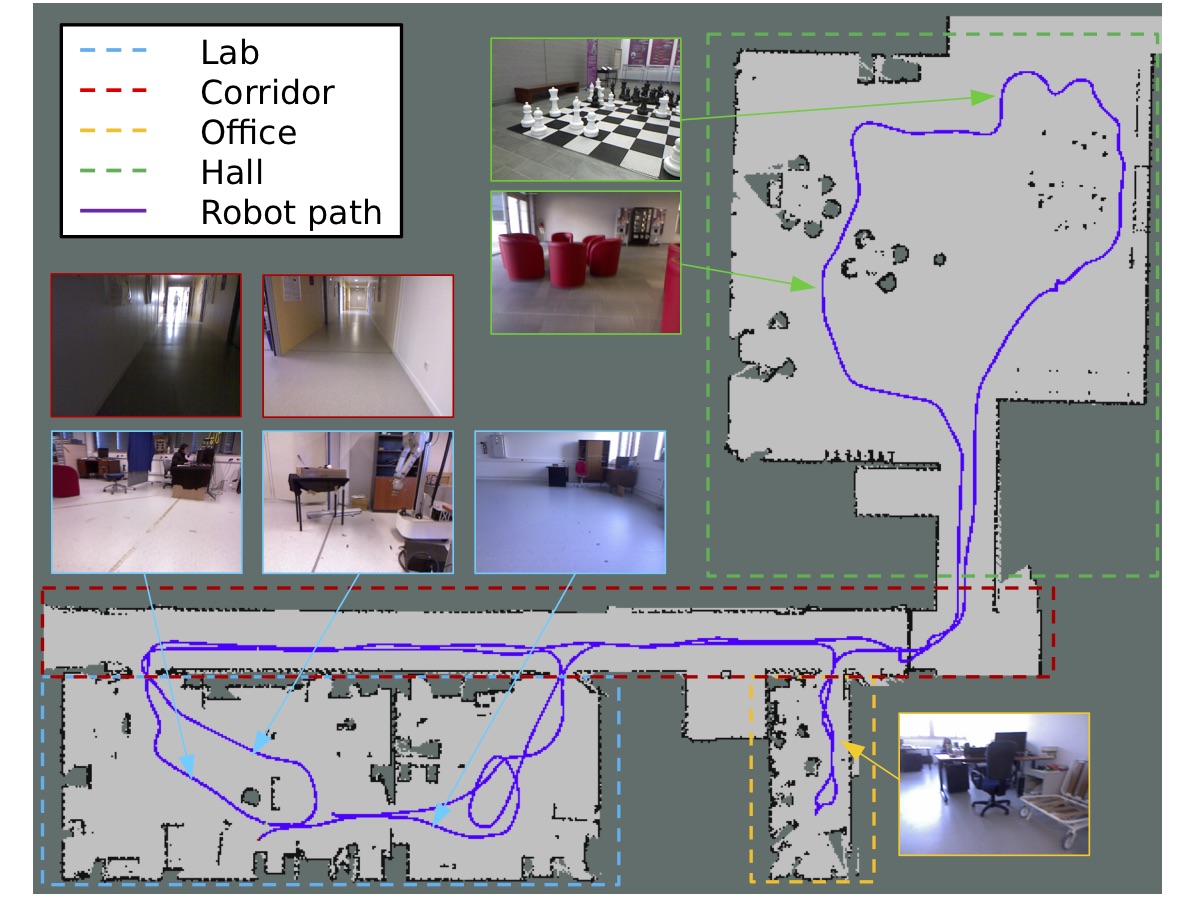}
       \caption{Map of the ENSTA building recorded in the \textit{ENSTA dataset}.}
       \label{fig:lab_regions}
\end{figure*}

The first dataset (denoted here as the \textit{ENSTA dataset}) was collected from a pioneer 3DX robot, with a Kinect RGB-D camera mounted at 1 meter from the ground and tilted slightly downward. The robot was equipped with a laser range finder, and a SLAM algorithm (Hector mapping~\cite{KohlbrecherMeyerStrykKlingaufFlexibleSlamSystem2011}) was used to simultaneously localize the robot in its environment and produce the occupancy grid. To build the dataset, we manually controlled the robot in an office building in order to visit corridors, laboratory, hall and offices. Within this environment, the robot was typically observing chairs, desks, or boxes. We recorded a 15 minutes length video sequence at 5Hz with the robot moving at a 0.5m/s average speed, in which a large variety of views and lightning conditions were captured (See Figure~\ref{fig:lab_regions}). In total, around 4000 RGB-D images were collected this way, associated with the position in the occupancy grid where they were taken.

The second dataset was constructed from the publicly available \textit{Washington dataset}, and more specifically, from the \textit{RGB-D scenes dataset} \cite{lai2011large}. This dataset is composed of 8 video sequences of indoor scenes with everyday-life objects placed on tabletops. In total, around 1500 RGB-D frames are available in this dataset along with bounding boxes around objects. \celine{However, these annotations are not well-suited for object-based saliency evaluation, as they consist in bounding boxes around a limited list of objects (other are just considered as distractors). For saliency, we rather need pixelwise annotations for every object of the scene.}

The third dataset is the \textit{Ciptadi dataset}~\cite{ciptadi2013depth}, designed for RGB-D saliency evaluation. This dataset was recorded on a mobile robot in a house, and contains everyday-life objects and scenes. In total 80 RGB-D frames are available, with a pixelwise saliency annotation. \celine{This dataset is used for two purposes: first, to confirm that our saliency technique is able to generalize from a dataset to another one. Second, to validate the performance of our method with annotations are not produced by our team, thus enhancing the reliability of our experiments.}

To evaluate the performance of our saliency technique, we manually labeled 100 randomly chosen images from the \textit{ENSTA dataset} and \textit{RGB-D scenes}. \celine{Annotations were done to be consistent with our definition of saliency. As a recall, we defined saliency objects as elements standing on planar surfaces (either floor or table), that can be detected by our segmentation technique. Note that this can be contradictory with bottom-up saliency, e.g. red plugs on a white wall will not be salient and white furniture in front of a white wall will be salient.} Annotations are such that we have a ground truth masks and a list of bounding boxes around salient objects. Those frames were removed from the dataset and used for evaluation only.

To evaluate the saliency model, we analyze the final performance reached by the classifier when all samples of the training set are used. We denote in this section our incremental saliency learning approach and produced saliency maps as ISL.

% \begin{figure*}
%       \includegraphics[height = 22cm]{mosaic_results.pdf}
%       \caption{Sample results of our saliency and bounding boxes proposal approaches (green highlights). \textbf{RGB}: the color input image. \textbf{GT}: the manually segmented ground truth. \textbf{Seg.}: the automatic object segmentation based on depth map. \textbf{ISL}: saliency map obtained from our incremental saliency learning (ISL) approach. \textbf{BMS}: BMS~\cite{zhang2013saliency} saliency map. \textbf{CAM}: CAM~\cite{zhou2015cnnlocalization} saliency map, \textbf{Vocus}: VOCUS2~\cite{frintrop2015traditional} saliency map. \textbf{EB+SB}: top 5 bounding boxes proposal obtained with the SegBoxes (yellow) and the re-ranked EdgeBoxes (blue). \textbf{EB}: top 5 bounding boxes proposal obtained with EdgeBoxes.}
%       \label{fig:sampleResults}
% \end{figure*}

\subsubsection{Depth segmentation vs. RGB saliency}
We first demonstrate the capacity of the saliency model to generalize what was learned from the depth segmentation, and produce a reliable estimation of the saliency on the whole image. This generalization is made possible by two factors: first salient objects often show common visual properties. The classifier is able to find those properties and is therefore able, to a certain extent, to find salient objects that were not detected by the object detector. Second, the datasets are such that the same objects are visible at different point of views. This way, if the object detector fails at identifying an object for a given frame, this object may be detected for other point of views. The classifier then extrapolates those point of views and is able to retrieve undetected salient objects.

\begin{figure*}
\centering
      \includegraphics[width = 15cm]{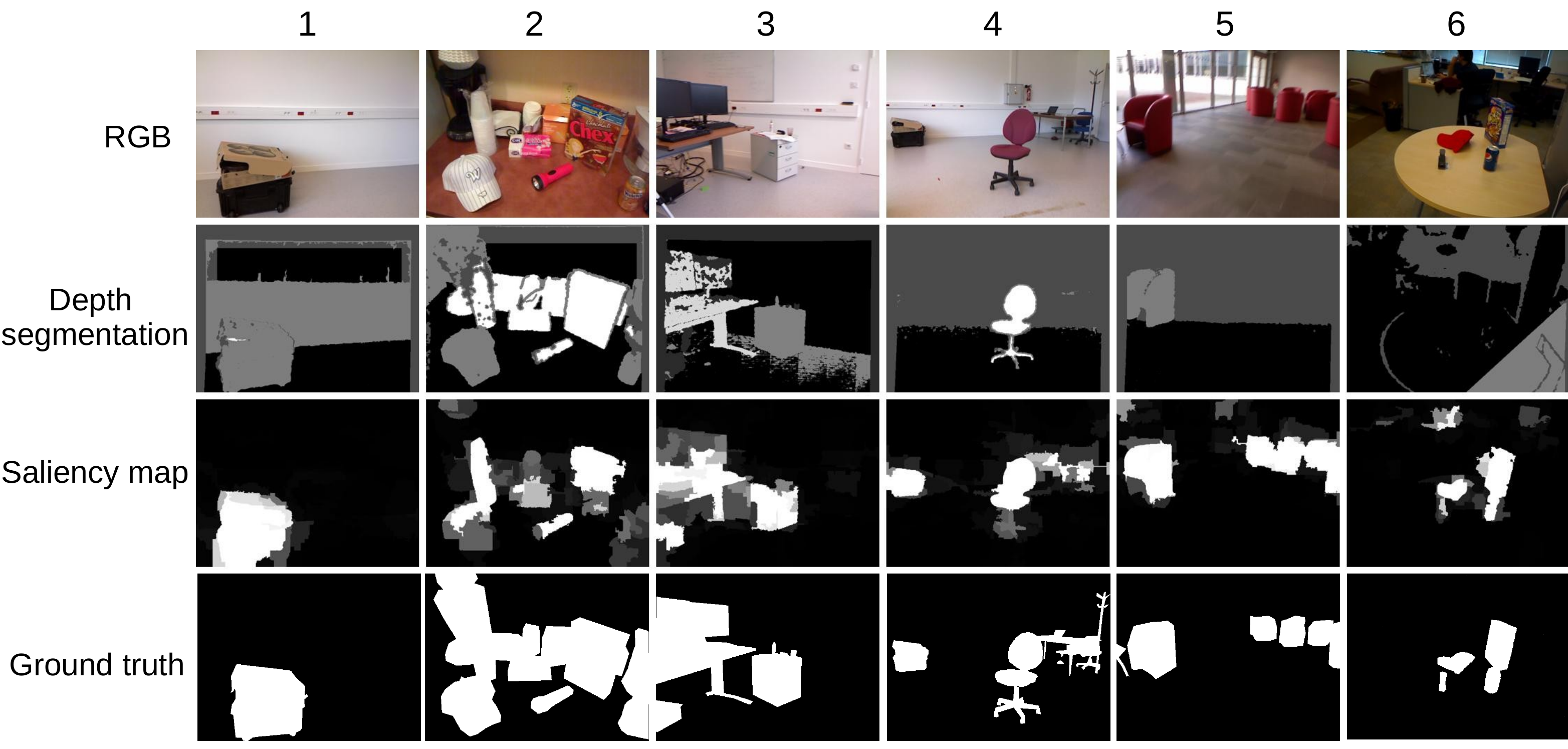}
      \caption{Generalization capabilities of the classifier. Compared with the ground truth, the depth segmentation mask is reliable, but partial. Conversely, saliency maps provide an estimation for each pixel of the input image.}
      \label{fig:generalizationCapacity}
\end{figure*}

Figure~\ref{fig:generalizationCapacity}, second row, shows a set of segmentation results. Recall that for a segmentation mask, black and white pixels represent \labelname{not salient} and \labelname{salient} portions of the image. Grey pixels of the segmentation mask represent areas where the algorithm could not clearly determine the state of saliency. Except cases where the segmentation fails (sample 6) because of a bad plane estimation, the segmentation mask produces a pretty reliable saliency segmentation. However, this segmentation is only partial because of the many undetermined areas, thus making the incremental learning of saliency (third row) useful to recover missing data. For example, the segmentation algorithm is such that nothing salient can be detected further than 4 meters away (samples 4, 5), but saliency estimation is applied on the whole image and the generalization capability of the classifier makes it possible to detect salient objects further than four meters. Second, reflective surfaces are often hard to detect by the Kinect sensor (computer screen on sample 3, black plastic on sample 2). However, the aspect of salient reflective objects can be partially learned and fully retrieved based on the RGB data only. Third, the segmentation algorithm is very restrictive and is often not able to detect salient elements if they are in contact with a border of the image (samples 1, 2,5) or badly clustered by the segmentation (mobile container of sample 3 mixed with the floor). Conversely, the saliency algorithm provides an estimate of saliency even if the object is partially cut, occluded, or captured with a poor image quality.

\subsubsection{Saliency map accuracy}

\celine{To demonstrate the accuracy of our saliency model, we selected three publicly available saliency algorithms and computed the ROC curves for each method on each of the three datasets. First, \textbf{SALGAN}~\cite{Pan_2017_SalGAN} is among the most accurate RGB saliency methods according to the \textit{MIT saliency benchmark}~\cite{mit-saliency-benchmark}. Second, we use the the \textbf{DSS} algorithm~\cite{2016arXiv160301976L} that produces an object-oriented kind of saliency and is one of the best performing method on the object-based ECSSD benchmark~\cite{shi2016hierarchical}. Third, we compare our method with saliency maps produced with the \textbf{CAM}~\cite{zhou2015cnnlocalization} model. This model is trained to detect objects among the 1000 classes of ILSVRC. For a fair comparison, we disabled classes that were not present in the images of our datasets (i.e. their output score were systematically set to 0), so that the produced saliency maps were responsive to relevant objects only. In addition, the maps produced by the CAM approach have the same low resolution than our model. We therefore apply the superpixels approach presented in Section \ref{sec:saliencyEstimation} to increase the resolution of these maps.}

To evaluate our feature extractor versus the one proposed in previous work~\cite{craye2016RL-IAC}, we generate saliency maps from both the CNN-based feature extractor (denoted as \textbf{ISL} here), and the former feature extractor (denoted as \textbf{ISL-Make3D}). Last we evaluate the performance of the \textbf{segmentation}. As saliency is not estimated on the whole image in this case, we replace pixels labeled as \labelname{unknown} or \labelname{unavailable} by a \labelname{not salient} label. For the Ciptadi dataset, Ciptadi \textit{et al.} have proposed a set of saliency maps for comparison on this dataset. We then present Ciptadi's results on this dataset only.

\begin{figure*}
   \centering
   \begin{tabular}{ccc} 

      \includegraphics[height = 4.5cm]{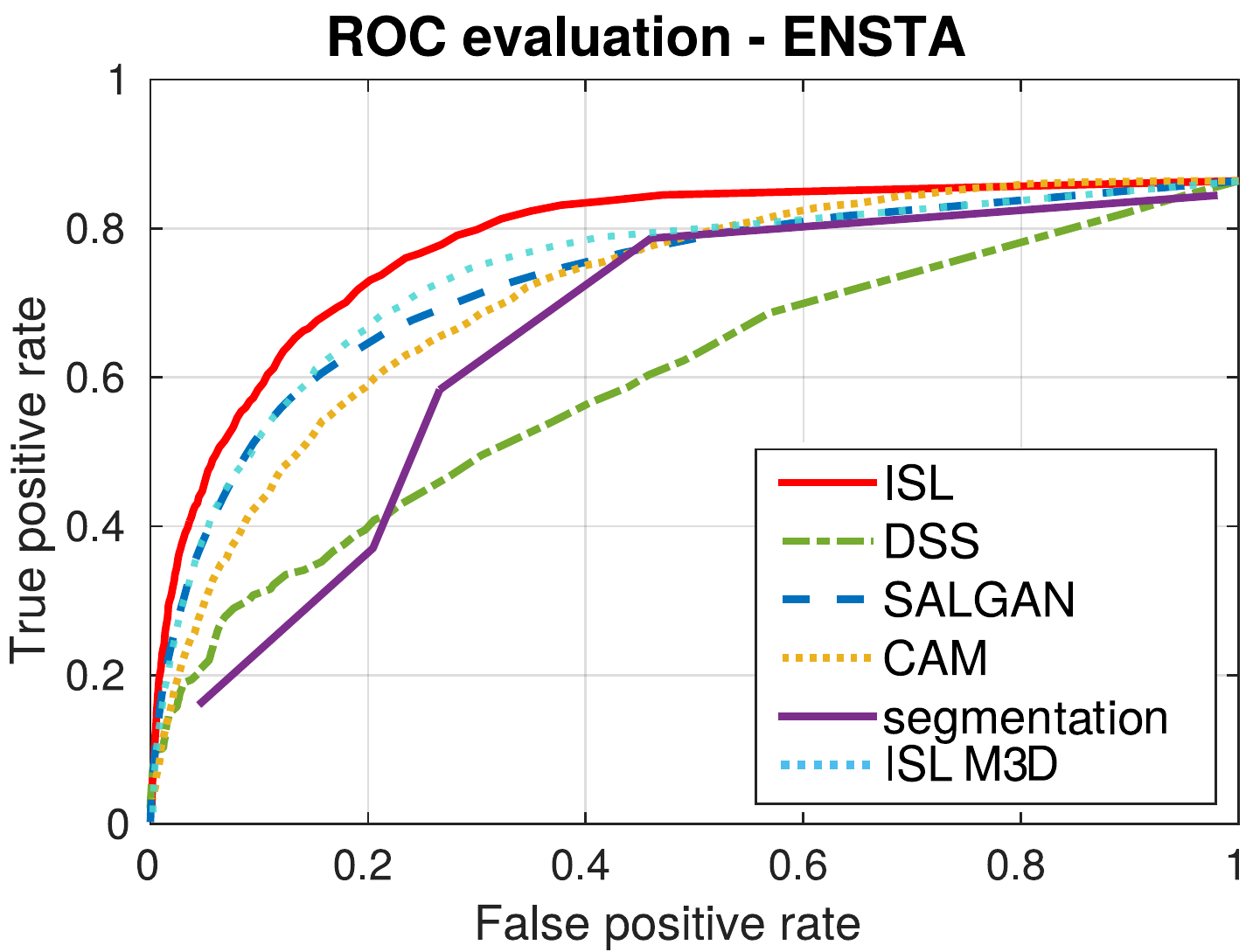}&
      \includegraphics[height = 4.5cm]{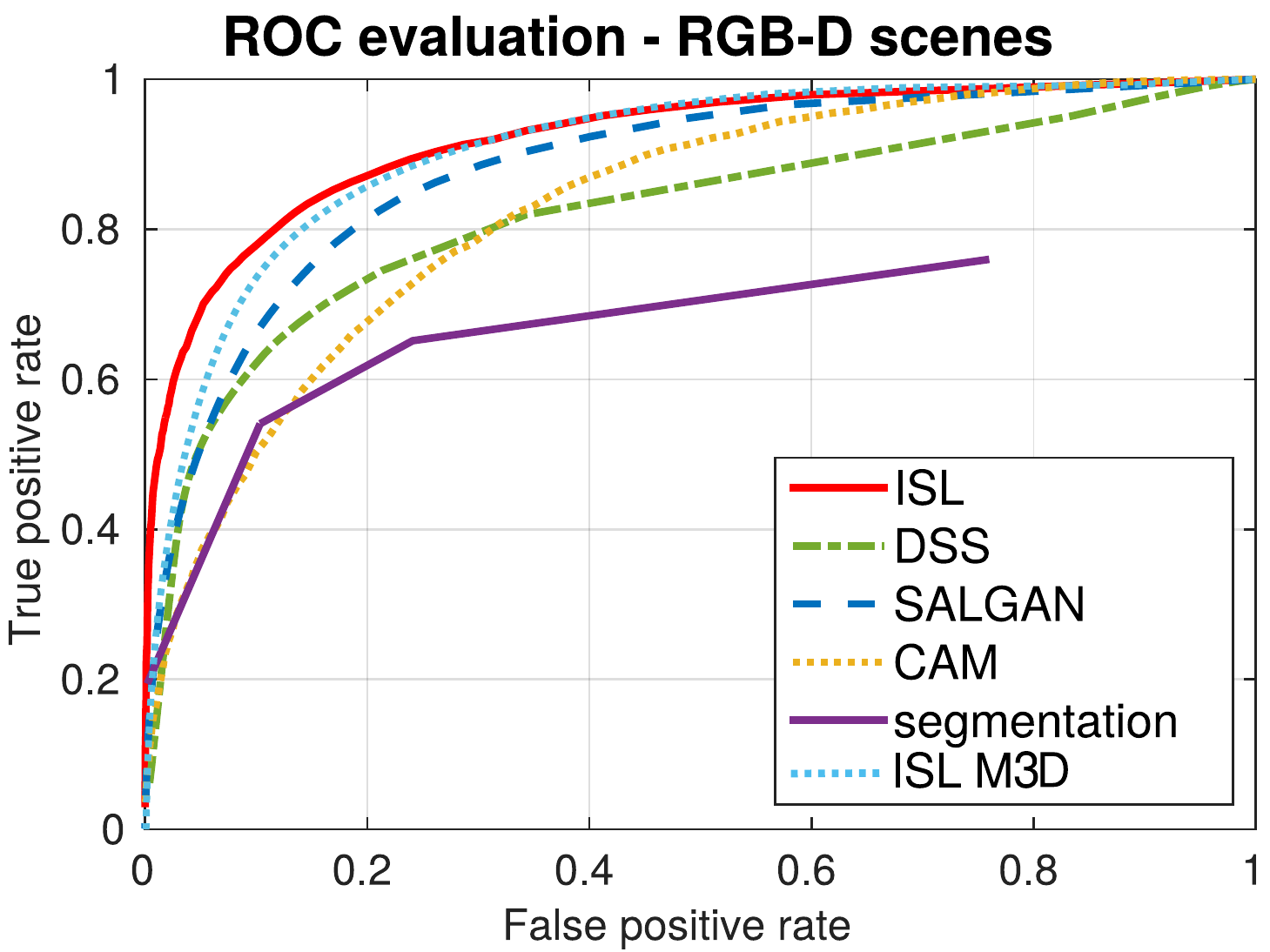}&
      \includegraphics[height = 4.5cm]{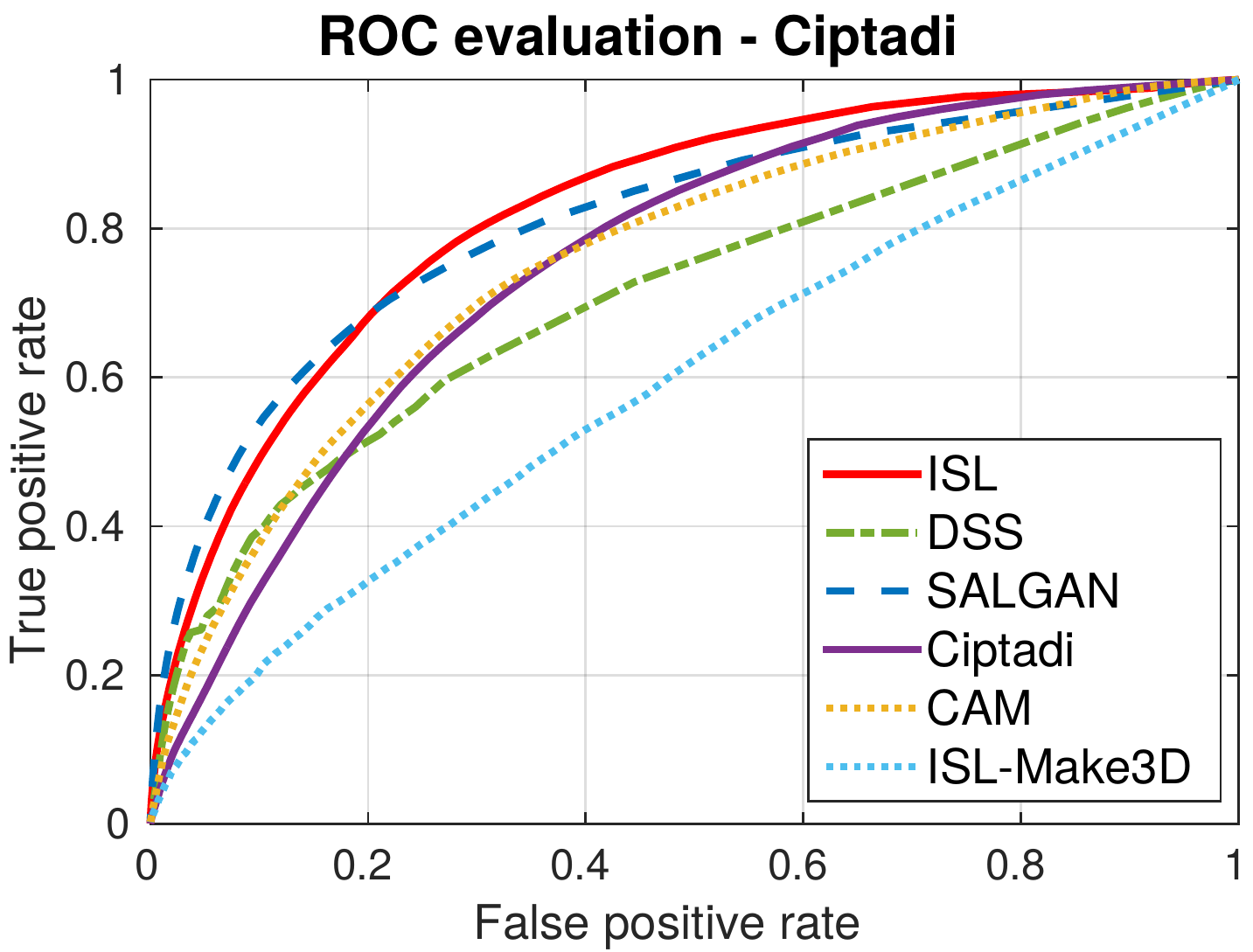}\\
   \end{tabular} 
      \caption{ROC curves of several saliency approaches on three different datasets}
      \label{fig:saliencyROC}
\end{figure*}

The results of the ROC-based evaluation are reported in Figure~\ref{fig:saliencyROC} and suggest that ISL significantly outperforms the evaluated bottom-up and top-down techniques on both \textit{ENSTA} and \textit{RGB-D scenes} datasets. Although slightly below ISL, our technique trained with the Make3D features is still performing well and confirms this trend. These result were expected on those two datasets, because our model is trained from a learning signal that is close to the ground truth, in a specific environment. \celine{Surprisingly, DSS is performing quite poorly for an object-based state-of-the-art saliency. We believe that is is because DSS tends to enhance the most saliency element mostly. As the ECSSD mainly consist of images with a single or few salient objects, this feature is not penalized in this dataset. However, our datasets are more challenging and usually contain several salient objects at the same time, thus making DSS less efficient.}

To demonstrate that the trained model is also usable in other kinds of environments, we use the Ciptadi dataset. \celine{The Ciptadi dataset has its own annotations on a pixelwise basis, so that results can be objectively compared with state-of-the-art.} For this dataset, we obtain our results by training ISL on the whole RGB-D scenes dataset, and creating saliency maps with this model. This time again, ISL is the best performing approach, but the Make3D version has a very poor performance. This result is explained by the very good generalization capacity of the CNN types of features.

We also observe that all of the evaluated techniques outperform the depth segmentation. This is because the ROC curve is estimated on the whole image, while segmentation only returns a partial saliency estimation. As missing information is replaced by \labelname{not salient} labels, the produced ROC curve has a low true positive rate.

\setlength{\tabcolsep}{2pt}
\newcolumntype{M}[1]{>{\centering\arraybackslash}m{#1}}

\begin{figure*}
\centering
      \includegraphics[width = 16cm]{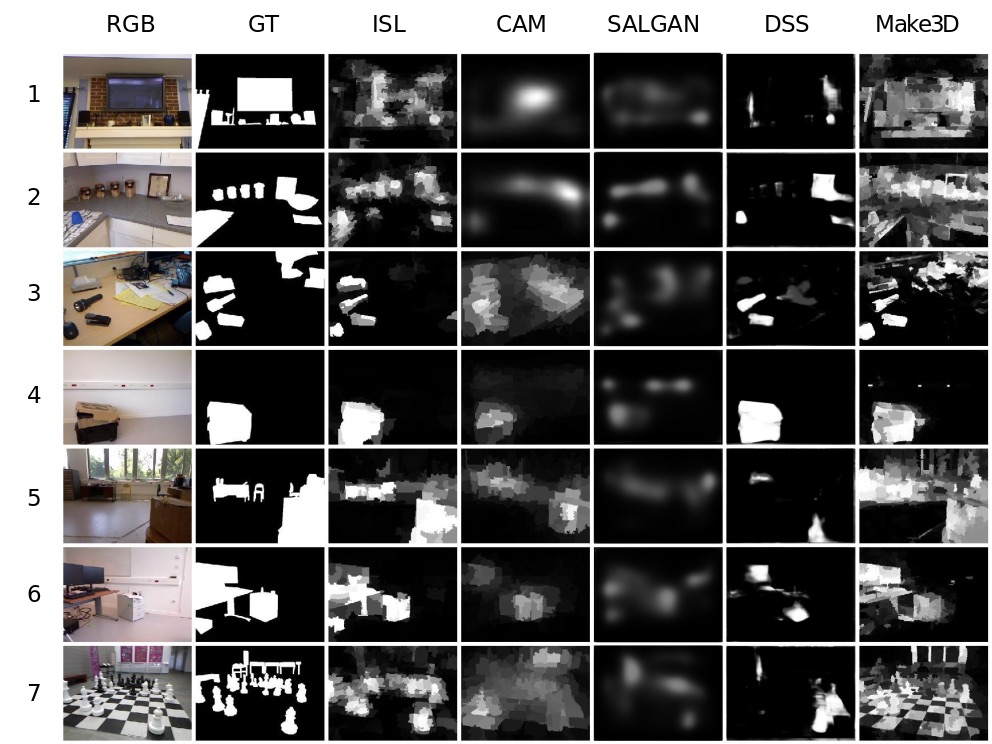}
      \caption{Comparison of several state of the art approaches with ISL, on Ciptadi (sample 1 and 2), RGB-D scenes (sample 3) and ENSTA (samples 4, 5, 6 and 7) datasets}
      \label{fig:stateOfTheArt}
\end{figure*}

In figure \ref{fig:stateOfTheArt}, a visual comparison between the saliency maps is presented. Samples 1 and 2 are from the Ciptadi dataset. For these samples, the results provided by ISL do not look as neat as in the other datasets. This is because ISL is used in an environment it was not trained for. Regarding the performance of ISL for the other samples, the superpixel reconstruction approach makes it possible to retrieve shapes of salient objects (in spite of the low-resolution feature maps produced by the output of the CNN). When applied to the CAM saliency map, the superpixel reconstruction does not provide such good results. This might be because the produced saliency is much more diffuse (as a comparison, the CAM algorithm is displayed samples 1 and 2 without superpixel reconstruction). \celine{As explained earlier, we notice that DSS tends to enhance only a few salient objects or portion of objects, thus leading to a lot of false negatives in the evaluation}. Second, ISL is learned from a segmentation derived from a depth map. This way, salient and not salient elements are determined from geometrical criteria rather than from RGB textures. As a results, ISL avoids the detection of distractors such as windows or trees outside (sample 5), or red power outlet (samples 4, 6), that are visually salient but irrelevant for an indoor mobile robot. Lastly, it enhances elements that are not naturally salient (mobile container on sample 6) but consistent with our definition. The saliency maps produced by the Make3D features have a better capacity for retrieving fine details of objects than ISL (samples 2, 6), as this type of feature extraction does not decrease the original resolution. However, ISL based on deep features have a much better generalization capability. Considering samples 1 and 2, where training was done on another dataset, or the chessboard of sample 6, where salient elements (the pawns) are of the same color than the ground (black and white squares), ISL clearly provides a much better saliency estimation.

\subsubsection{Bounding box proposals}
\begin{figure*}
   %\centering
   \begin{tabular}{cc} 

      \includegraphics[height = 5.7cm]{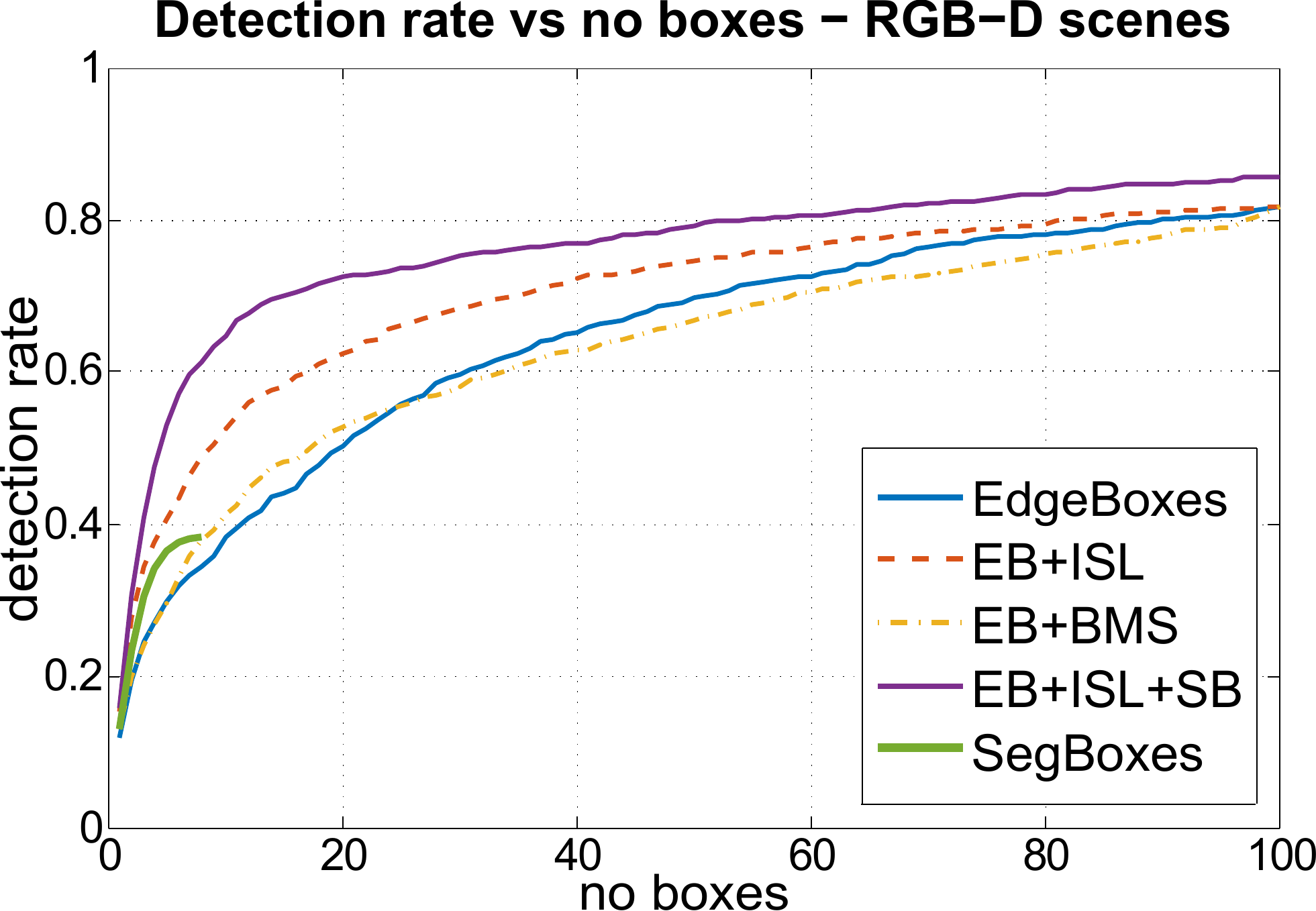}&
      \includegraphics[height = 5.7cm]{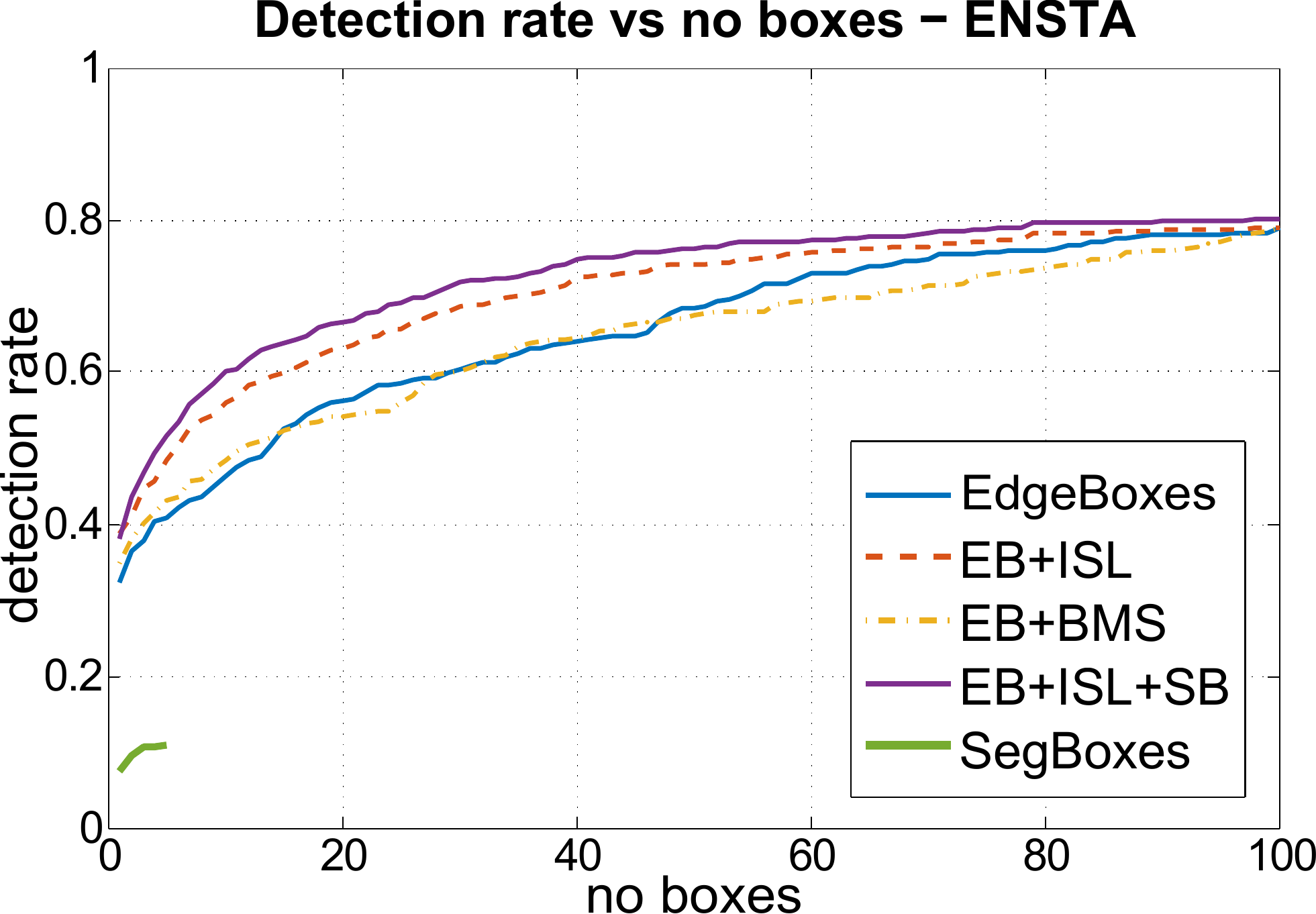}\\
(a) & (b)
   \end{tabular} 
      \caption{Detection rate vs number of boxes proposal comparison on the \textit{RGB-D scenes} dataset (a) and ENSTA dataset (b).}
      \label{fig:detectionrate}
\end{figure*}

\begin{figure*}
      \centering
      \includegraphics[width = 14cm]{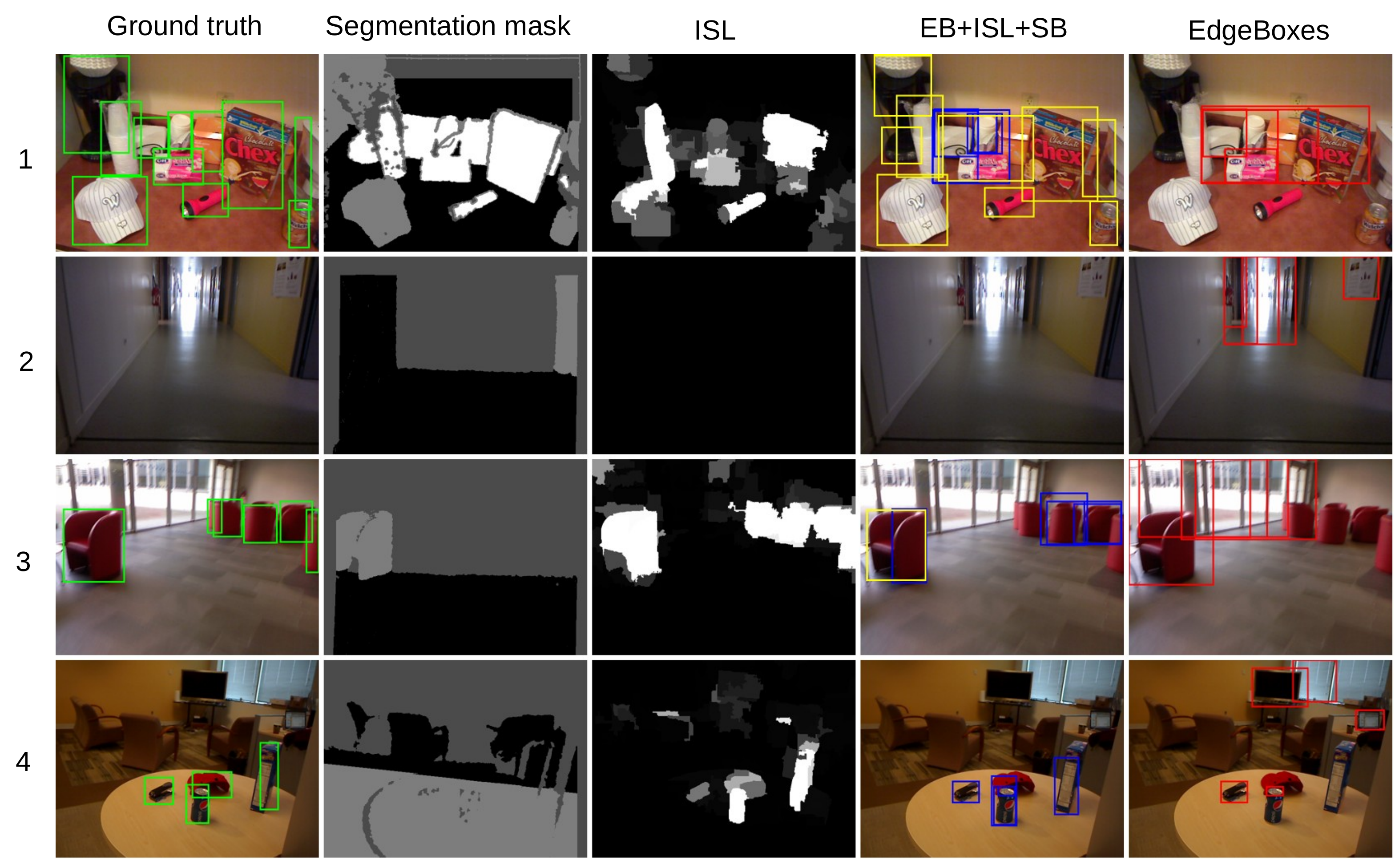}\\
      \caption{Sample results of bounding box proposals versus EdgeBoxes}
      \label{fig:bboxesMosaic}
\end{figure*}

We now demonstrate that ISL can be used to produce relevant bounding boxes around objects. To this end, we run the EdgeBoxes~\cite{zitnick2014edge} algorithm for each frame of the \textit{ENSTA} and \textit{RGB-D scenes} evaluation sets and keep the 100 best ranked bounding boxes along with their $h_b^{in}$ scores.  These boxes are used as a reference to evaluate our method. Then, these EdgeBoxes are re-ranked based on the saliency map to make boxes containing salient pixels better ranked than others. For that, we rank each of these boxes according to the SCscore defined in Section~\ref{sec:bboxes}. To demonstrate the ability of our saliency map to produce relevant box proposals, we calculate an SCscore based on ISL saliency maps, and another one based on BMS~\cite{zhang2013saliency} saliency maps (denoted as \textbf{EB+ISL} and \textbf{EB+BMS} in Figure \ref{fig:detectionrate}). Lastly, we generate the SegBoxes from the depth segmentation process, as described in Section \ref{sec:bboxes}. We also produce an SCscore for each of them. We filter out Segboxes having a low SCscore (below 0.2 in our case). Those SegBoxes, obtained from depth segmentation are complementary to the RGB-based EdgeBoxes and allow the detection of additional relevant boxes. The remaining SegBoxes are reported as \textbf{SegBoxes} in Figure \ref{fig:detectionrate}. In practice, a small number of SegBoxes are detected on each frame (between 0 and 7 on average). Lastly, we combine the re-ranked EdgeBoxes and the SegBoxes to produce a better set of box proposals. This approach is presented in Figure \ref{fig:detectionrate} and Figure~\ref{fig:bboxesMosaic} as \textbf{EB+ISL+SB}.

The evaluation metric is the detection rate versus the number of proposal, based on the \textit{intersection over union} measure (IoU=0.5 here) to count the number of detections. This measure is used by Zitnick \textit{et al.}~\cite{zitnick2014edge} to evaluate their performance over state of the art approaches. To obtain the detection rate for $N$ proposals, we consider the $N$ best ranked box proposals and measure the proportion of boxes in the ground truth that have an IoU score over 0.5 with at least one of the proposals.

Numerical results are reported in Figure \ref{fig:detectionrate} for both  \textit{ENSTA} and \textit{RGB-D scenes} datasets. As expected, the use of ISL maps to improve the EdgeBoxes ranking allows a much better detection rate on both datasets. Moreover, using a bottom-up saliency map such as BMS instead of ISL does not show significant improvements on both datasets. The SegBoxes usually propose relevant candidates, possibly not detected by the EdgeBoxes. Because they are complementary to the EdgeBoxes, combining the two approaches significantly improve the detection rate on both datasets. However, the number of proposals is low (between 0 and 7 most of the time), and they do not cover the entire image as they are produced from the depth segmentation.

\newcolumntype{M}[1]{>{\centering\arraybackslash}m{#1}}

Figure \ref{fig:bboxesMosaic} shows sample results of the top 5 EdgeBoxes (column \textbf{EdgeBoxes}), top 5 EbdgeBoxes re-ranked by the SCscore with ISL (displayed in column \textbf{EB+ISL+SB}, blue boxes), and Segboxes (same column, yellow boxes). The SegBoxes almost always provide relevant boxes, but many objects are also missed this way, either because they are too far to be segmented (sample 3), or because segmentation failed (sample 4). In this case, the remaining objects locations are recovered by the EdgeBoxes. Again, the use of ISL to rank the EdgeBoxes favors boxes that surround salient elements while removing distractors such as windows (sample 3). Lastly, it is possible to cope with frames that do not contain any salient object (sample 2) by filtering boxes with an SCscore below a certain threshold (0.01 in our case).

\subsection{RL-IAC}

\subsubsection{Experimental setups}
A strong limitation when using robots, especially in scenarios involving online learning, is the reproducibility of the experiments. If a single experiment requires the robot to explore a building for hours, the total number of possible trials is rather limited, and the efficiency of a method may be hard to analyze. For that reason, we rely on semi-simulated setups to run a large number of experiments in parallel without any particular user monitoring. The semi-simulated setups are created from the recorded sequences of the \textit{ENSTA} and \textit{RGB-D scenes datasets}. We call them semi-simulated as saliency is learned from real images taken from these sequences, but actions taken by the robot are simulated.

The first setup is constructed from the \textit{ENSTA dataset}. From this sequence, we build before our experiments a navigation graph based on the technique described in Section~\ref{sec:regions} (although map building and saliency learning could be run simultaneously). The navigation graph used in all our experiments is the one represented in Figure~\ref{fig:navigationGraph}. Proto-regions were arbitrarily defined to be of 5 meters length. We then considered the positions of each observation recorded during the sequence, and we associated each of them to a region. When selecting a position in a given region, we select one among all associated frames, we consider the position of this frame, and we simulate the displacement of the robot to reach it. Once attained, we use this observation to update our model. To get an overview of the incremental map building and the simulated robot displacement, a video is available on the project's webpage\footnote{\url{https://github.com/cececr/RL-IAC}}.

\begin{figure*}
   \centering
   \begin{tabular}{cc}
   		Short corridor building & Long corridor building\\
        \includegraphics[height = 7cm]{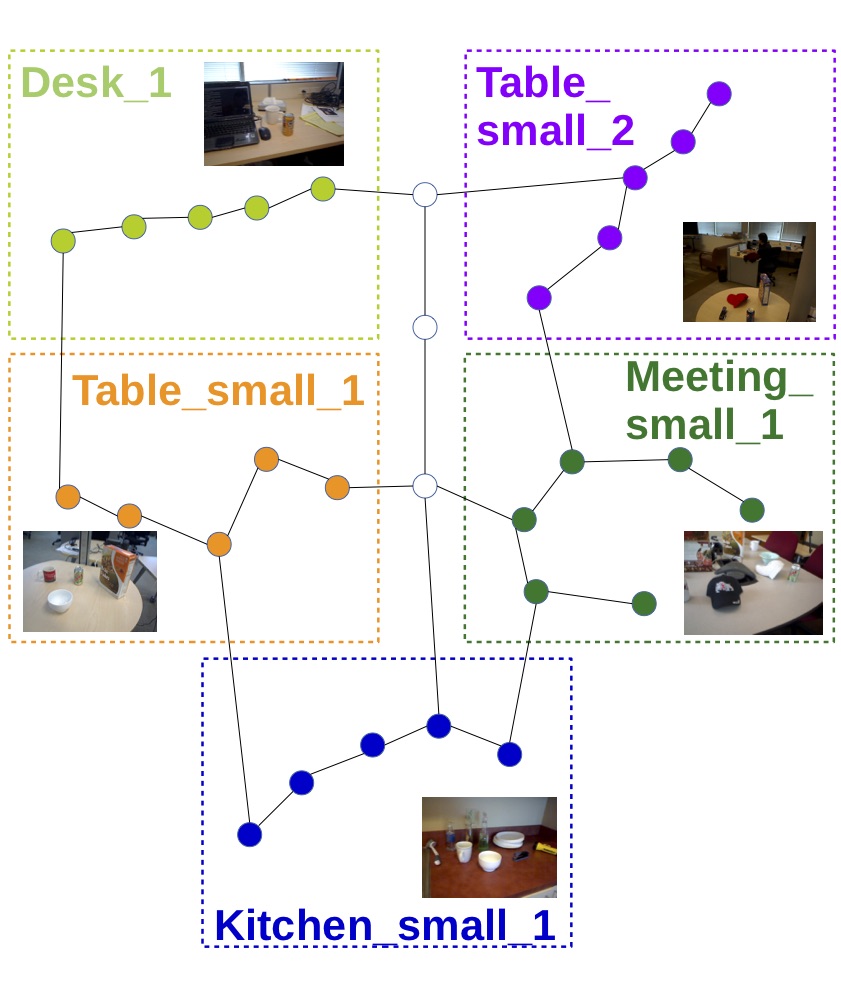} & 
       \includegraphics[height = 7cm]{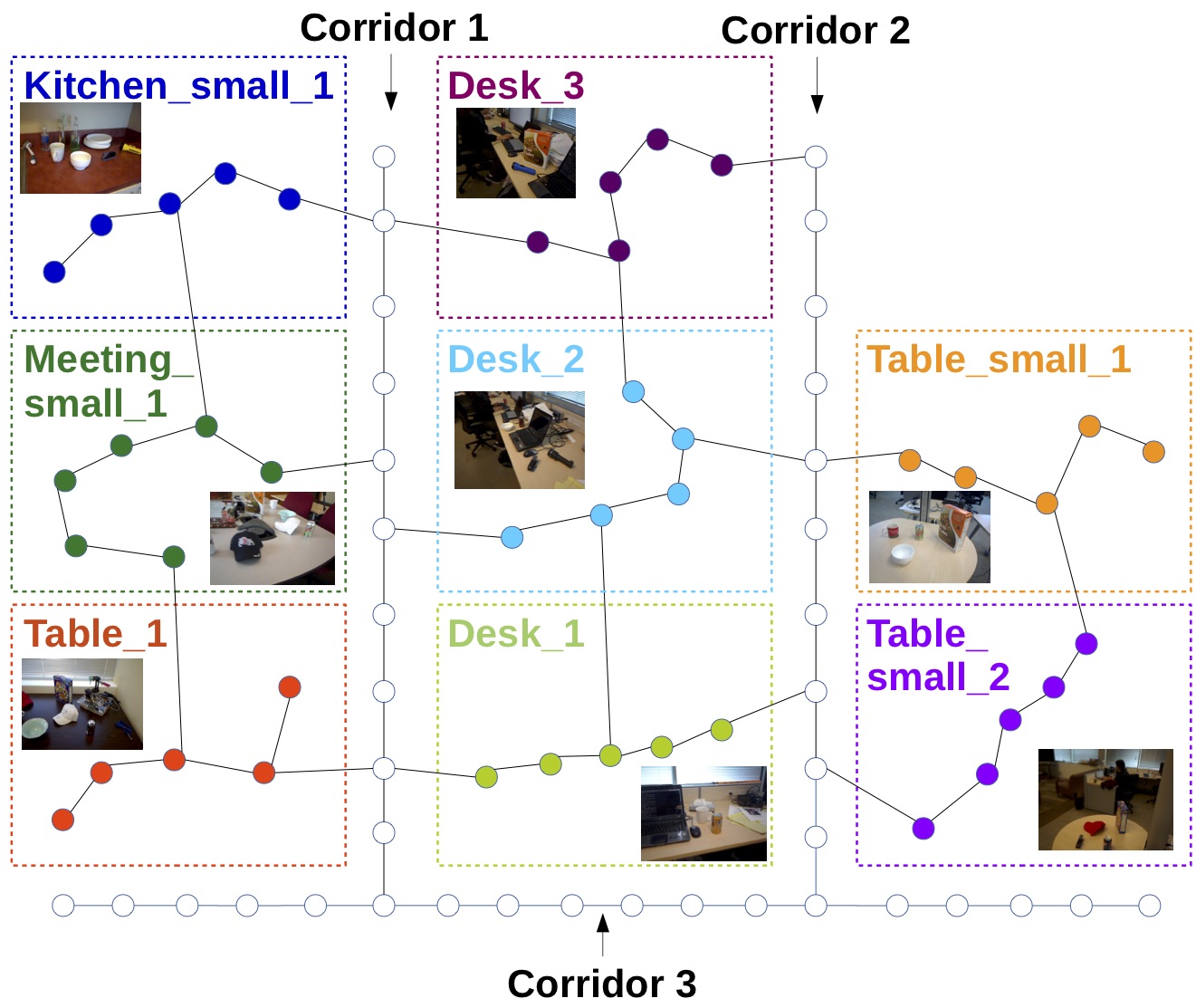}
       \end{tabular}
      \caption{\textit{short corridor} and \textit{long corridor} artificial buildings created from the \textit{RGB-D scene dataset}}
      \label{fig:rgbdScenesBuilding}
\end{figure*}

The two other datasets consist in artificial buildings constructed from the \textit{RGB-D scene dataset}. Each of the eight video sequences of the dataset is recorded in a single particular room (kitchen, office, meeting room), so that our artificial building contains rooms (one for each sequence) divided into 5 to 6 regions, with some regions connected to other rooms (as if there were doors and corridors between rooms). We created two different building configurations, illustrated by Figure~\ref{fig:rgbdScenesBuilding}. The first artificial building, denoted as the \textit{short corridor building}, is composed of five of the video sequences, and contains a short corridor of three regions to switch between rooms. The second one, denoted as the \textit{long corridor building}, is composed of the eight video sequences and contains three long corridors. To construct the navigation graph in each room, we cut each of the sequence into five or six sub-sequences of equal length, and we created an arbitrary trajectory to travel across the sequences. We also limited the number of connections per region to four. This corresponds to the four possible actions the robot can take, namely \labelname{up}, \labelname{down}, \labelname{left} and \labelname{right}.

To simulate the displacements of the robot, we considered the following sequence of steps:
\begin{enumerate}
\item The robot grabs an RGB-D frame, extracts features and segmentation, and updates the meta-learner;
\item the robot determines the next region to visit given learning progress and Q-learning training;
\item the robot determines the next position to reach in this region;
\item the robot moves to this position;
\item while moving, the robot updates the learner based on the previous RGB-D frame;
\item before taking the next RGB-D frame, the robot waits for the displacement and the learner update to be both finished.
\end{enumerate}

In our experiments, this sequence was repeated 3000 times. Each estimated error rate was timestamped with the simulated time starting at the beginning of the experiment. This timestamps was then used to plot our results. To obtain the simulated time, we measured for each iteration the time spent by the system to compute steps 1 to 3 (not simulated), and we added the longest step between steps 4 and 5, as they are supposed to be run in parallel and wait for the other to be finished. Table~\ref{tab:RLIACExecutionTime} provides additional measurements to get a better overview of the execution time for each step.

\renewcommand{\arraystretch}{1.5}
\newcolumntype{M}[1]{>{\centering}m{#1}}
\begin{table}
\centering
\begin{tabular}{M{3.5cm}M{1.7cm}M{1.7cm}}
  \hline
  \textbf{Processing} & \textbf{Min time} & \textbf{Max time}  \tabularnewline 
  \hline\hline
  Meta-learner update & 100 ms & 150 ms \tabularnewline 
  Q-learning training & 250 ms & 300 ms \tabularnewline 
  Learner update & 23 ms & 13.5 s \tabularnewline
  Robot displacement & 0 ms & 22 s \tabularnewline
  \hline
  
\end{tabular}
\caption{Min and max processing time for the main steps of RL-IAC}
\label{tab:RLIACExecutionTime}
\end{table}

\begin{figure*}
\centering
\includegraphics[height = 3.3cm]{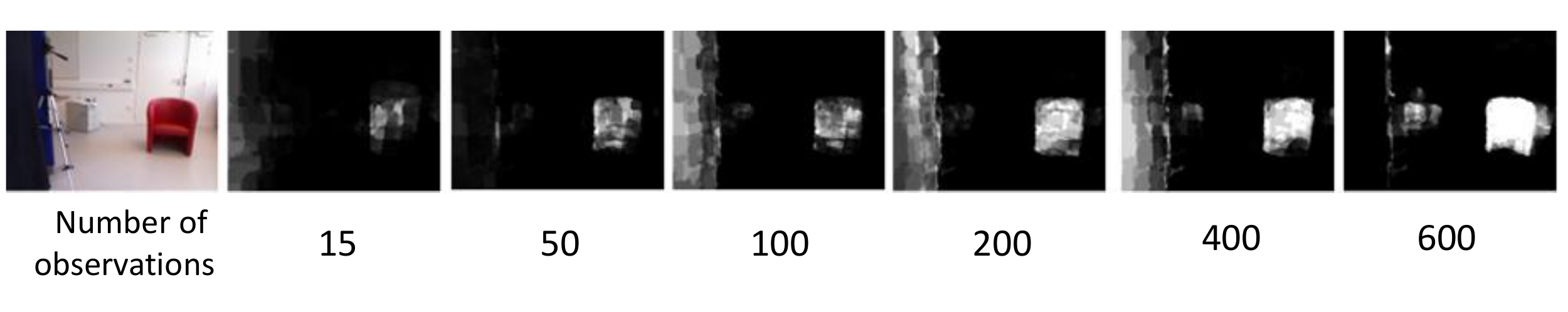}
      \caption{Evolution of the saliency as number of observation increases. In this example, the frame were learned in chronological order, and the seat was observed for the first time only after 400 frames.}
      \label{fig:saliencyEvolution}
\end{figure*}

To simulate the robot's displacement, we considered an average speed of 0.5 m/s, and we measure the time for a robot to reach a certain point by considering the euclidean distance to this point and a constant speed of 0.5 m/s. The maximum displacement time is then bounded by the distance between two adjacent regions. To get a rough idea of the simulated time for a single experiment, the five steps of an iteration take on average 10 seconds. Given the 3000 successive iterations, an experiment then lasts for 8 hours.

\subsubsection{Evolution of the saliency}

We first look at the evolution of the saliency quality during incremental learning. Figure \ref{fig:saliencyEvolution} first shows a qualitative example of the evolution of the saliency at a given point of view, while the sequence is used in chronological order for training the classifier. We can observe the generalization capability because even before the seat was observed (in frame 400), the classifier is already able to recognize it as a salient element, because it has already learned a partial model of the background. For a better visualization of this evolution, a video is also available online on the project's webpage.

\subsubsection{Exploration efficiency}
To demonstrate the benefits of exploring the environment using RL-IAC, we now compare the evolution of the saliency with different exploration strategies on the three datasets. In a previous work~\cite{craye2016RL-IAC}, we demonstrated that RL-IAC was outperforming IAC's action policy. We here investigate other types of explorations.

\newcolumntype{M}[1]{>{\centering\arraybackslash}m{#1}}
\begin{figure*}[t]
   \centering
   \begin{tabular}{ccc} 

      \includegraphics[height = 4.9cm]{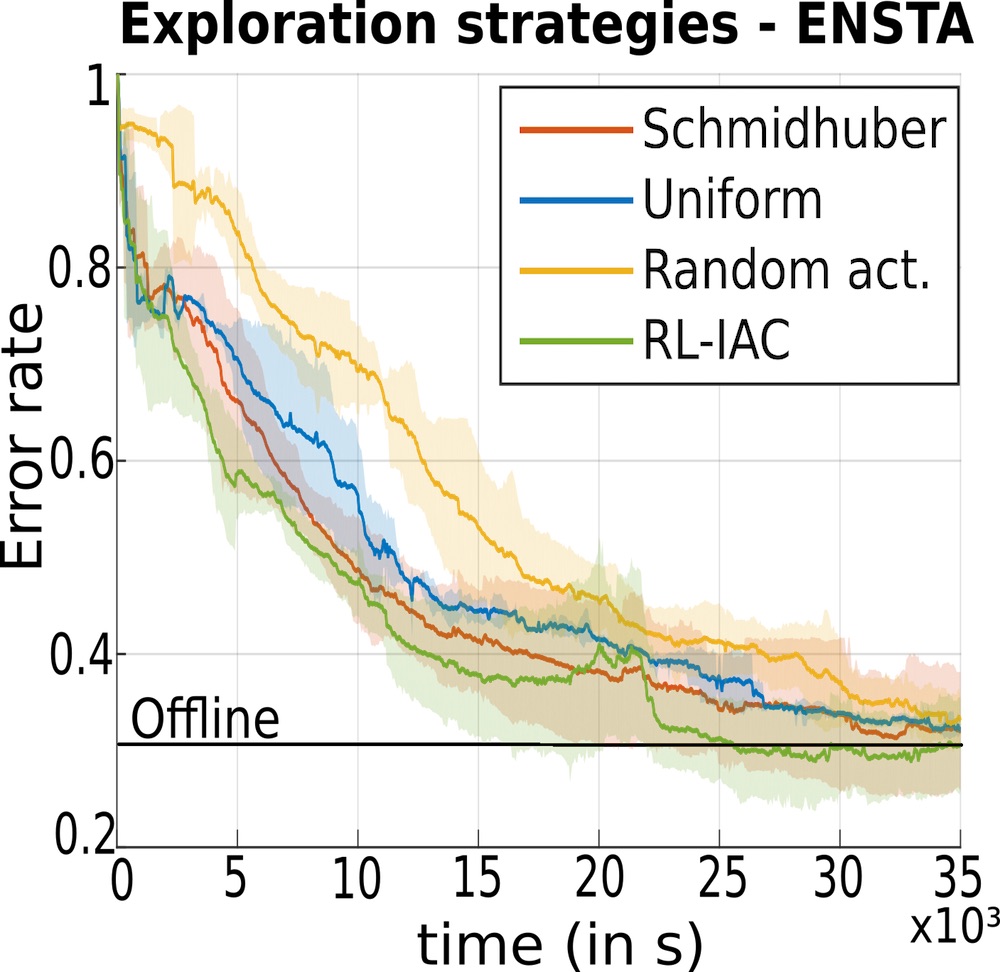}~~&
      \includegraphics[height = 5cm]{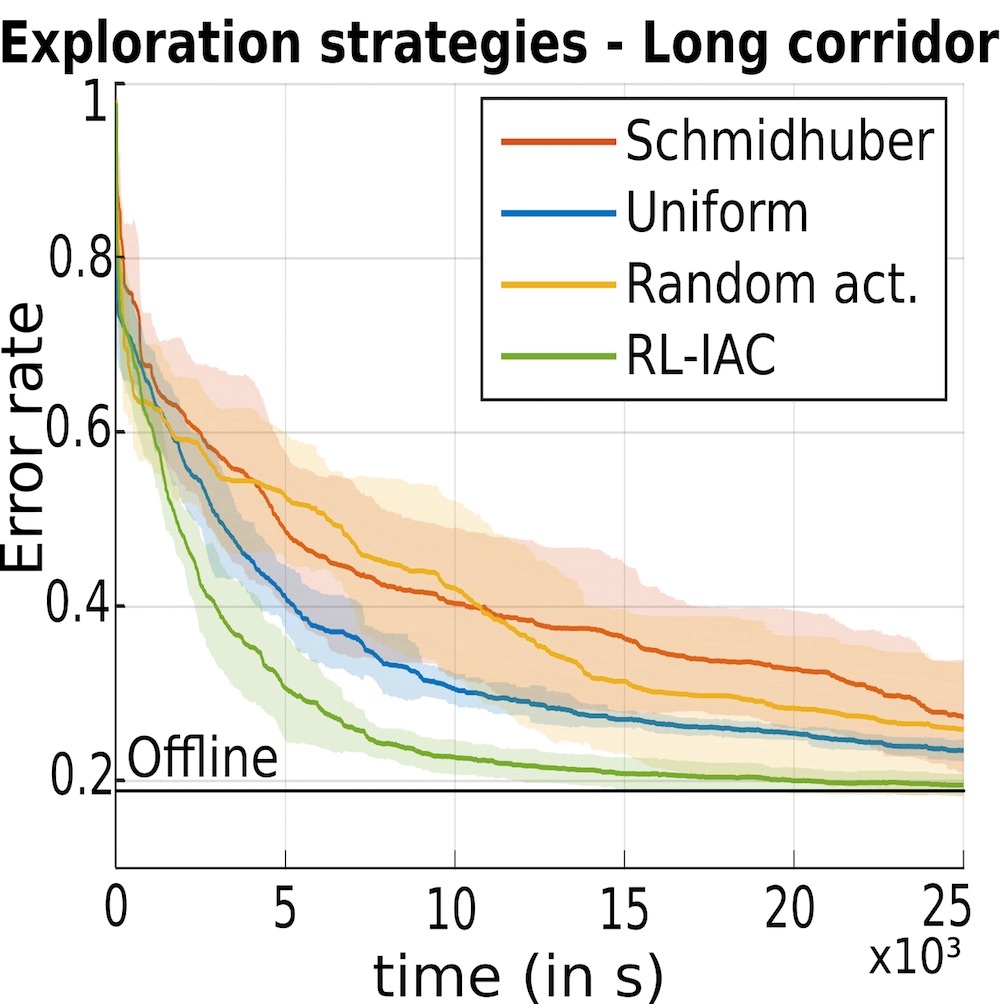}~~&
      \includegraphics[height = 5cm]{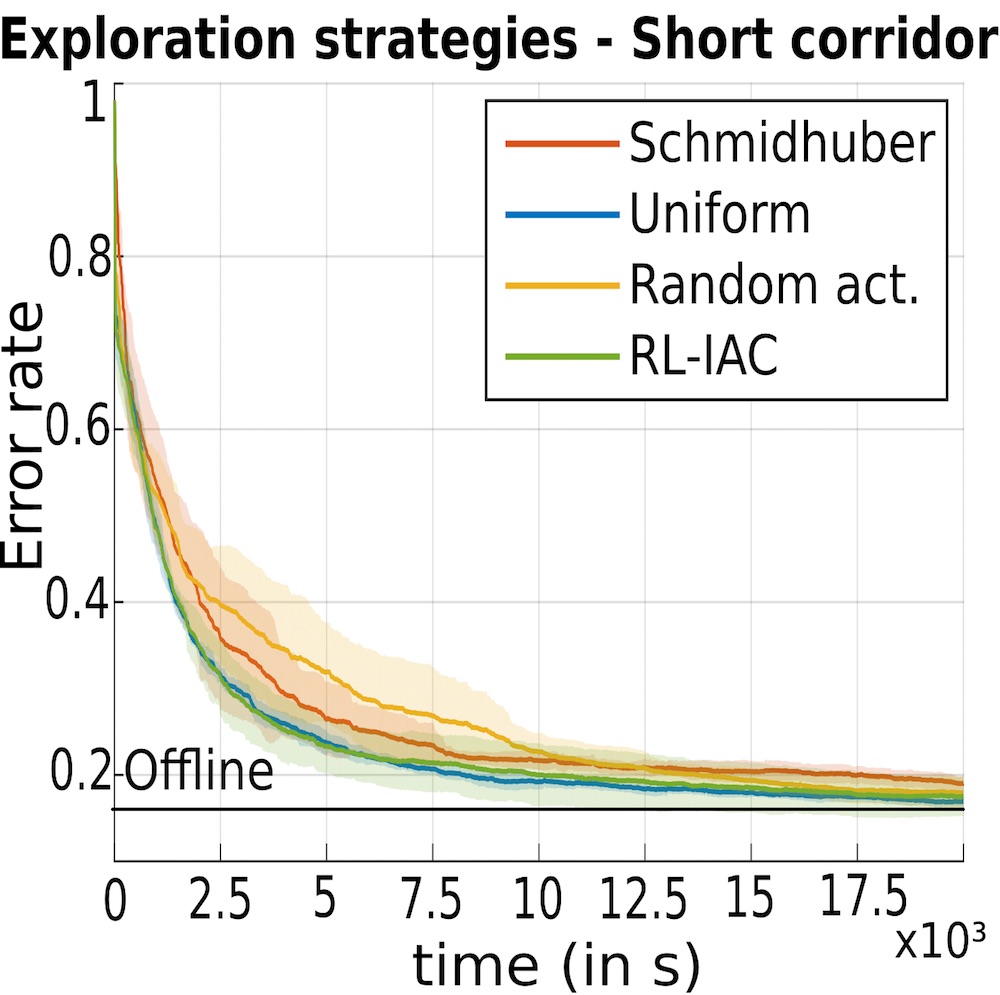}\\
(a) & (b) & (c)
   \end{tabular} 
      \caption{Error rate evolution for several exploration strategies on three different environments}
      \label{fig:RLIACExplorationStratgy}
\end{figure*}

In mobile robotics navigation, the goal is generally to have a good coverage of the environment to explore so as to get an accurate mapping. Our goal is not to make a mapping of the environment, but using an exploration based on an extensive and efficient coverage of the environment is a good baseline to compare with. For this reason, our first exploration strategy consists in determining an exploration pattern covering the whole regions of the environment, and repeating this pattern until the end of the experiment. In Osswald \textit{et al.}, pre-defined map and navigation graph is used as a prior for exploration. An efficient exploration route is obtained by solving a traveling salesman problem (TSP) in their map, based on the Concorde software~\cite{applegate2006concorde}. Similarly, we then used Concorde with our regions configurations, to find an optimal map coverage that is to be used in our experiments.

A few approaches rely on learning progress to guide exploration in a reinforcement learning context. In particular, Schmidhuber~\cite{schmidhuber1991curious} or Lopes \textit{et al.}~\cite{lopes2012exploration} have used Q-learning to guide the robot's displacements in a context where the only reward is the learning progress. This kind of approach is very similar to RL-IAC, but differs at a critical point: while a single Q-learning is run during the experiment and directly decides the next action of the robot in Schmidhuber's approach, we define and solve a new problem with Q-learning after each robot's displacements. We then use the entire problem to find the next best displacement rather than following a policy from a partially trained Q-matrix. Our second approach to compare with is then following Schmidhuber's approach: instead of running virtual displacement simulations to train our Q-matrix, we run a single update of the Q-matrix after arriving in a given region. We also use an $\epsilon$-greedy approach (50\% random) to decide the next action to take.

Each exploration strategy was tested 10 times on each dataset and results are reported based on the average and variance over those experiments. The performance of the system was evaluated using the evolution of the \textit{overall error rate} of the system: based on the reference frames on which a ground truth is available, we compare the estimated saliency map for all of these frames with the available ground truth. We then use the formula provided by Equation \ref{eq:error_rate} on each frame and take the average error. Note that the \textit{overall error rate} is an extrinsic metrics used to evaluate the performance of the system. It then differs from the \textit{region error rate}, the intrinsic metrics (based on segmentation rather than ground truth) used to get an estimate of the error in each region.

Figure \ref{fig:RLIACExplorationStratgy} shows the evolution of the \textit{overall error rate} in time on both environments, for the 4 exploration strategies: 
\begin{itemize}
\item \textbf{RL-IAC}: As described in Section~\ref{sec:RL-IAC}. Selects the next region to visit from the Q-matrix, and the next position to reach in that region randomly.
\item \textbf{Uniform}: We drive the exploration by a uniform coverage of the environment, from the sequence of regions determined with the TSP heuristic. This pattern is repeated until the end of the experiment. The next region to visit is determined from the sequence, and the next position to reach in that region is taken randomly.
\item \textbf{Schmidhuber}: Similar to RL-IAC, except that the Q-matrix is updated after each observation rather than running a batch of simulations.
\item \textbf{Random act.}: To get a worse case scenario, we select a random action to reach a region, and random position in that region.
\end{itemize}

On all datasets, RL-IAC is the method with the fastest decreasing error. The uniform exploration has a reasonable performance, even similar to RL-IAC in the short-corridor dataset. This can be explained by the fact that this setup only has a very small number of uninformative regions. RL-IAC, by evaluating progress, is precisely efficient at detecting such uninformative regions. This is even more visible in the \textit{large corridor} experiment, where almost 50\% of the regions are part of the corridors, which are typically uninformative. Schmidhuber has a varying performance depending on the dataset. We actually found this approach very sensitive to the parameters of the experiment, and performing well with very different parameters than RL-IAC. For example, to converge rapidly enough, a large percentage of random actions were necessary (typically 50\%), while RL-IAC only works with 10\% of random actions. Lastly, and as expected, the random action is providing the worse performance, sometimes close to Schmidhuber's approach. \celine{As a comparison, we also plot the error rate of the model trained offline (constant in time). The offline version performs roughly the same as the online one when enough samples have been acquired. However, the main difference is that the online version is flexible to changes in the environment, while the offline is not.}

\subsubsection{Time allocation in the environment}

\begin{figure*}
\centering
    \begin{tabular}{cc} 
	\includegraphics[height=5cm]{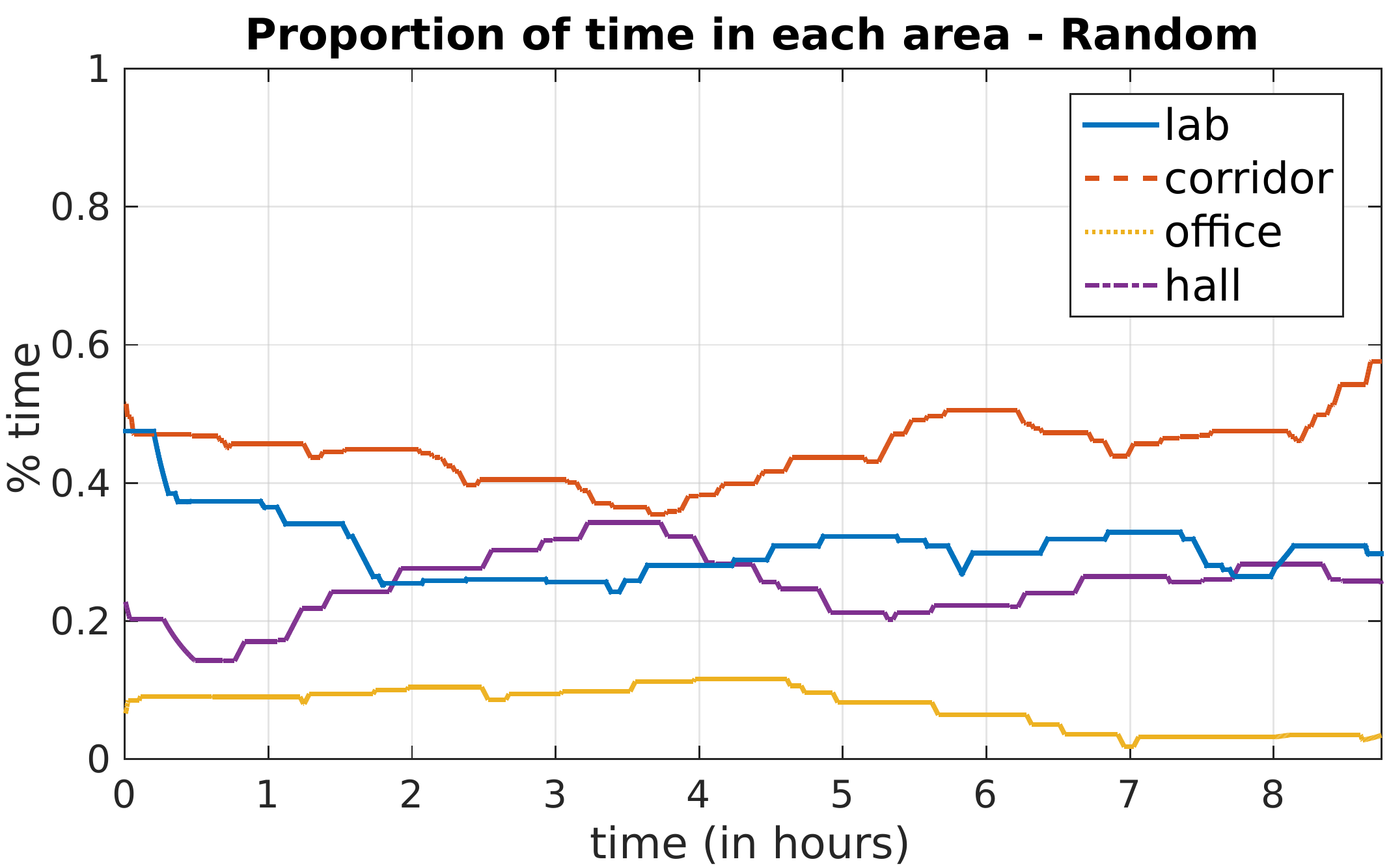} & 
	\includegraphics[height=5cm]{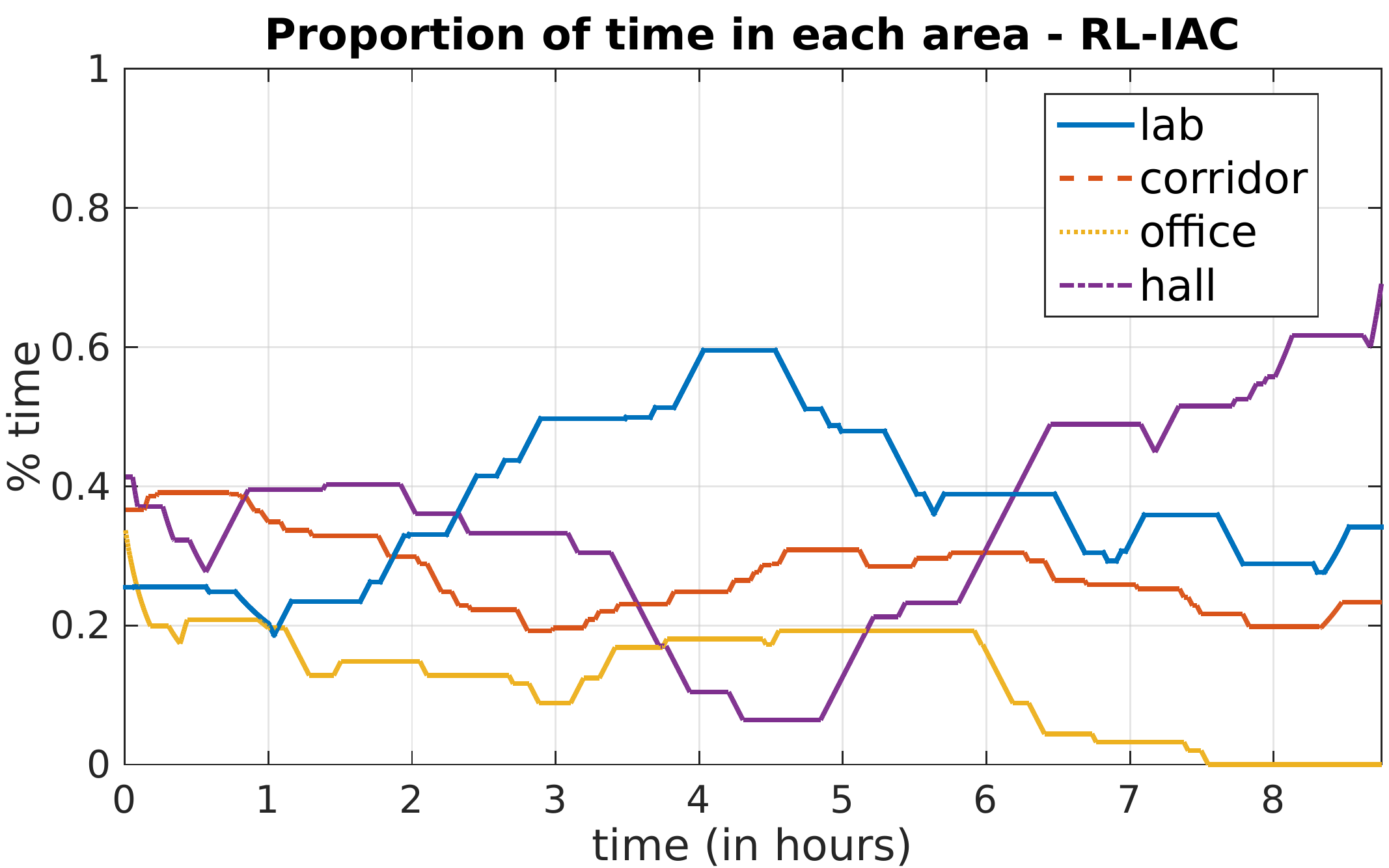}\\
	(a) & (b) \\ 
	
    \end{tabular}
    \caption{Time spend for (a) random position selection (b) RL-IAC}
    \label{fig:perCluster}
\end{figure*}

To get a better insight of the way exploration is done by the robot, we divide the building in 4 main areas, namely \textit{lab}, \textit{office}, \textit{corridor} and \textit{hall} (See Figure \ref{fig:lab_regions}). In these areas the difficulty to learn saliency is not the same. For example, the corridor does not contain any salient element, whereas the hall is a very large room with many salient items and many distractors. We compare in Figure \ref{fig:perCluster} the average percentage of time spent in each area when using RL-IAC and when using random (\textbf{Random act.}) exploration strategies. The graphs have been constructed by using a sliding widow of 500s over the whole experiment, and measuring for each window the number of frames obtained in each area. For random exploration, the time spent in each area is roughly the same all along the sequence and proportional to each area size. 40\% to 50\% of the time is spent in the corridor, whereas 10\% is spent in the office. With RL-IAC, the time spent in the corridor (the least \say{interesting} area) oscillates between 30\% and 20\%, except at the beginning, and almost 20\% is spent in the office. Moreover, the time spent in exploring each area is evolving in time: the time spent in the office finally decreases to 0\%, because no progresses are made in there anymore. In the middle of the sequence, most of the time is spent exploring the lab, while most of the time is spent in the hall at the end of the exploration.

\section{Discussion}

\celine{We took the assumption that salient elements are objects. This is of course a restrictive case, but this assumption holds in many indoor applications. Any definition of saliency could be used to replace the one chosen here, as soon as an appropriate learning signal can be used to learn this kind of saliency. For example, in~\cite{craye18}, we propose another kind of learning signal and another kind of saliency, that is provided by a foveated platform.}

\celine{Whatever the saliency definition is, our approach will depend on the quality of the learning signal. In the current work, this information is of good quality at short range, but very partial as it does not give information close to the image borders and at long range. It would be interesting to study what are the ideal characteristics of this learning signal, for example if a lower quality but more complete signal would be relevant, or if improving the current segmentation quality would lead to a noticeable final performance increase.}

\celine{ In this paper, the robot is only exploring its environment and thus takes all decision in order to improve its saliency model. Such situation would be rare in real-world scenarios, but our approach can be easily integrated into a robot that has other tasks to fulfill as there is no theoretical problems in mixing intrinsic and extrinsic motivations~\cite{chentanez2004intrinsically}. This would result in a robot opportunistically exploring to improve its saliency model, for example taking a route through a less known area while going to its charging station.}

\section{Conclusion}
\label{sec:Conclusion}

In this article, we have presented a full architecture for learning to localize objects within a robot's exploration in an incremental and autonomous way. On the one hand, we described the main mechanisms for learning a model of visual saliency from a depth-based learning signal, and how to exploit this saliency model to general bounding boxes around salient objects of the scene. On the other hand, we investigated how the robot could methodically explore its environment to learn the saliency model faster and better. We proposed the RL-IAC approach to guide exploration in that regard, by finding the best compromise between robot's displacement and learning. We have carried out several experimentation to demonstrate the accuracy of our saliency maps as compared with other state-of-the-art approaches, and the efficiency of our exploration technique.

\celine{A critical aspect that should be consider in future work is the use of an end-to-end deep learning framework that would both produce saliency and bounding box proposals. We so far separate feature extraction, feature combination, and bounding boxes generation, but deep learning offers a way to integrate all these components at the same time. This could take the form of a fully convolutional network that would produce both saliency and boxes. This could for example resemble the SSD architecture~\cite{liu2016ssd}. Additionally, neural networks are by essence online classifiers, which may be better suited than the proposed method based on random forests. Although incremental learning with deep neural network is still at an early stage, we could simplify the problem by only fine-tuning a small part of the network. Alternatively,
various concise CNN models such as binary CNN could be used to increase the efficiency.}

Two other possible directions would be worth investigating. First, running the incremental map building and RL-IAC at the same time. This way, the navigation graph would be constructed from scratch, without any prior environment exploration. Second, we would like to carry out experiments in a non simulated setup to have a fully operational system.

\section*{Acknowledgment}

The authors would like to thank the INRIA Flowers team, and especially Pierre-Yves Oudeyer for the valuable help on the IAC aspect. 

\ifCLASSOPTIONcaptionsoff
  \newpage
\fi



%\bibliographystyle{IEEEtran}
\bibliographystyle{plain}
\bibliography{Report}

\begin{thebibliography}{10}

\bibitem{alexe2012measuring}
Bogdan Alexe, Thomas Deselaers, and Vittorio Ferrari.
\newblock Measuring the objectness of image windows.
\newblock {\em Pattern Analysis and Machine Intelligence, IEEE Transactions
  on}, 34(11):2189--2202, 2012.

\bibitem{ali2014contextual}
Haider Ali, Faisal Shafait, Eirini Giannakidou, Athena Vakali, Nadia Figueroa,
  Theodoros Varvadoukas, and Nikolaos Mavridis.
\newblock Contextual object category recognition for rgb-d scene labeling.
\newblock {\em Robotics and Autonomous Systems}, 62(2):241--256, 2014.

\bibitem{applegate2006concorde}
David Applegate, Ribert Bixby, Vasek Chvatal, and William Cook.
\newblock Concorde tsp solver, 2006.

\bibitem{baranes2009r}
Adrien Baran{\`e}s and P-Y Oudeyer.
\newblock R-iac: Robust intrinsically motivated exploration and active
  learning.
\newblock {\em Autonomous Mental Development, IEEE Transactions on},
  1(3):155--169, 2009.

\bibitem{barto1998reinforcement}
Andrew~G Barto.
\newblock {\em Reinforcement learning: An introduction}.
\newblock MIT press, 1998.

\bibitem{bazeille2011incremental}
St{\'e}phane Bazeille and David Filliat.
\newblock Incremental topo-metric slam using vision and robot odometry.
\newblock In {\em Robotics and Automation (ICRA), 2011 IEEE International
  Conference on}, pages 4067--4073. IEEE, 2011.

\bibitem{bjorkman2010active}
M{\aa}rten Bj{\"o}rkman and Danica Kragic.
\newblock Active 3d scene segmentation and detection of unknown objects.
\newblock In {\em Robotics and Automation (ICRA), 2010 IEEE International
  Conference on}, pages 3114--3120. IEEE, 2010.

\bibitem{borji2010online}
Ali Borji, Majid~Nili Ahmadabadi, Babak~Nadjar Araabi, and Mandana Hamidi.
\newblock Online learning of task-driven object-based visual attention control.
\newblock {\em Image and Vision Computing}, 28(7):1130--1145, 2010.

\bibitem{borji2013state}
Ali Borji and Laurent Itti.
\newblock State-of-the-art in visual attention modeling.
\newblock {\em Pattern Analysis and Machine Intelligence, IEEE Transactions
  on}, 35(1):185--207, 2013.

\bibitem{brafman2003r}
Ronen~I Brafman and Moshe Tennenholtz.
\newblock R-max-a general polynomial time algorithm for near-optimal
  reinforcement learning.
\newblock {\em The Journal of Machine Learning Research}, 3:213--231, 2003.

\bibitem{mit-saliency-benchmark}
Zoya Bylinskii, Tilke Judd, Fr{\'e}do Durand, Aude Oliva, and Antonio Torralba.
\newblock Mit saliency benchmark.
\newblock http://saliency.mit.edu/.

\bibitem{canas2008overt}
Jos{\'e}~M Ca{\~n}as, Marta~Mart{\'\i}nez de~la Casa, and Teodoro Gonz{\'a}lez.
\newblock An overt visual attention mechanism based on saliency dynamics.
\newblock {\em International Journal of Intelligent Computing in Medical
  Sciences \& Image Processing}, 2(2):93--100, 2008.

\bibitem{caron2014neural}
Louis-Charles Caron, David Filliat, and Alexander Gepperth.
\newblock Neural network fusion of color, depth and location for object
  instance recognition on a mobile robot.
\newblock In {\em Computer Vision-ECCV 2014 Workshops}, pages 791--805.
  Springer, 2014.

\bibitem{cheng2011global}
Ming-Ming Cheng, Guo-Xin Zhang, Niloy~J Mitra, Xiaolei Huang, and Shi-Min Hu.
\newblock Global contrast based salient region detection.
\newblock In {\em Computer Vision and Pattern Recognition (CVPR), 2011 IEEE
  Conference on}, pages 409--416. IEEE, 2011.

\bibitem{chentanez2004intrinsically}
Nuttapong Chentanez, Andrew~G Barto, and Satinder~P Singh.
\newblock Intrinsically motivated reinforcement learning.
\newblock In {\em Advances in neural information processing systems}, pages
  1281--1288, 2004.

\bibitem{ciptadi2013depth}
Arridhana Ciptadi, Tucker Hermans, and James~M Rehg.
\newblock An in depth view of saliency.
\newblock In {\em Eds: T. Burghardt, D. Damen, W. Mayol-Cuevas, M. Mirmehdi, In
  Proceedings of the British Machine Vision Conference (BMVC 2013)}, pages
  9--13, 2013.

\bibitem{craye2016use}
C{\'e}line Craye, David Filliat, and Jean-Fran{\c{c}}ois Goudou.
\newblock On the use of intrinsic motivation for visual saliency learning.
\newblock In {\em 2016 Joint IEEE International Conference on Development and
  Learning and Epigenetic Robotics (ICDL-EpiRob)}. IEEE, 2016.

\bibitem{craye18}
C{\'e}line Craye, David Filliat, and Jean-Fran{\c{c}}ois Goudou.
\newblock Biovision: a biomimetics platform for intrinsically motivated visual
  saliency learning.
\newblock {\em IEEE Transactions on Cognitive and Developmental Systems}, 2018.

\bibitem{craye2016RL-IAC}
Celine Craye, David Filliat, and JF~Goudou.
\newblock Rl-iac: An exploration policy for online saliency learning on an
  autonomous mobile robot.
\newblock In {\em Intelligent Robots and Systems (IROS), 2016 IEEE
  International Conference on}, 2016.

\bibitem{ecins2016cluttered}
Aleksandrs Ecins, Cornelia Ferm{\"u}ller, and Yiannis Aloimonos.
\newblock Cluttered scene segmentation using the symmetry constraint.
\newblock In {\em Robotics and Automation (ICRA), 2016 IEEE International
  Conference on}, pages 2271--2278. IEEE, 2016.

\bibitem{erdem2013visual}
Erkut Erdem and Aykut Erdem.
\newblock Visual saliency estimation by nonlinearly integrating features using
  region covariances.
\newblock {\em Journal of vision}, 13(4):11, 2013.

\bibitem{fischler1981random}
Martin~A Fischler and Robert~C Bolles.
\newblock Random sample consensus: a paradigm for model fitting with
  applications to image analysis and automated cartography.
\newblock {\em Communications of the ACM}, 24(6):381--395, 1981.

\bibitem{frintrop2006vocus}
Simone Frintrop.
\newblock {\em VOCUS: A visual attention system for object detection and
  goal-directed search}, volume 3899.
\newblock Springer, 2006.

\bibitem{frintrop2010computational}
Simone Frintrop, Erich Rome, and Henrik~I Christensen.
\newblock Computational visual attention systems and their cognitive
  foundations: A survey.
\newblock {\em ACM Transactions on Applied Perception (TAP)}, 7(1):6, 2010.

\bibitem{frintrop2015traditional}
Simone Frintrop, Thomas Werner, and Germ{\'a}n~Mart{\'\i}n Garc{\'\i}a.
\newblock Traditional saliency reloaded: A good old model in new shape.
\newblock In {\em Proceedings of the IEEE Conference on Computer Vision and
  Pattern Recognition}, pages 82--90, 2015.

\bibitem{garcia2015saliency}
Germ{\'a}n~M Garc{\'\i}a, Ekaterina Potapova, Thomas Werner, Michael Zillich,
  Markus Vincze, and Simone Frintrop.
\newblock Saliency-based object discovery on rgb-d data with a late-fusion
  approach.
\newblock In {\em 2015 IEEE International Conference on Robotics and Automation
  (ICRA)}, pages 1866--1873. IEEE, 2015.

\bibitem{hamker2005emergence}
Fred~H Hamker.
\newblock The emergence of attention by population-based inference and its role
  in distributed processing and cognitive control of vision.
\newblock {\em Computer Vision and Image Understanding}, 100(1):64--106, 2005.

\bibitem{hosang2016makes}
Jan Hosang, Rodrigo Benenson, Piotr Doll{\'a}r, and Bernt Schiele.
\newblock What makes for effective detection proposals?
\newblock {\em IEEE transactions on pattern analysis and machine intelligence},
  38(4):814--830, 2016.

\bibitem{2016arXiv161104849H}
Q.~{Hou}, M.-M. {Cheng}, X.-W. {Hu}, A.~{Borji}, Z.~{Tu}, and P.~{Torr}.
\newblock {Deeply supervised salient object detection with short connections}.
\newblock {\em ArXiv e-prints}, November 2016.

\bibitem{hou2012image}
Xiaodi Hou, Jonathan Harel, and Christof Koch.
\newblock Image signature: Highlighting sparse salient regions.
\newblock {\em Pattern Analysis and Machine Intelligence, IEEE Transactions
  on}, 34(1):194--201, 2012.

\bibitem{huang2002novelty}
Xiao Huang and John Weng.
\newblock Novelty and reinforcement learning in the value system of
  developmental robots.
\newblock 2002.

\bibitem{itti2001computational}
Laurent Itti and Christof Koch.
\newblock Computational modelling of visual attention.
\newblock {\em Nature reviews neuroscience}, 2(3):194--203, 2001.

\bibitem{itti1998model}
Laurent Itti, Christof Koch, and Ernst Niebur.
\newblock A model of saliency-based visual attention for rapid scene analysis.
\newblock {\em IEEE Transactions on pattern analysis and machine intelligence},
  20(11):1254--1259, 1998.

\bibitem{jebari2012combined}
Islem Jebari, St{\'e}phane Bazeille, and David Filliat.
\newblock Combined vision and frontier-based exploration strategies for
  semantic mapping.
\newblock In {\em Informatics in Control, Automation and Robotics}, pages
  237--244. Springer, 2012.

\bibitem{KohlbrecherMeyerStrykKlingaufFlexibleSlamSystem2011}
S.~Kohlbrecher, J.~Meyer, O.~von Stryk, and U.~Klingauf.
\newblock A flexible and scalable slam system with full 3d motion estimation.
\newblock In {\em Proc. IEEE International Symposium on Safety, Security and
  Rescue Robotics (SSRR)}. IEEE, November 2011.

\bibitem{kompella2012autonomous}
Varan Kompella, Matthew Luciw, Marijn Stollenga, Leo Pape, and Jurgen
  Schmidhuber.
\newblock Autonomous learning of abstractions using curiosity-driven modular
  incremental slow feature analysis.
\newblock In {\em Development and Learning and Epigenetic Robotics (ICDL), 2012
  IEEE International Conference on}, pages 1--8. IEEE, 2012.

\bibitem{kragic2009object}
Danica Kragic.
\newblock Object search and localization for an indoor mobile robot.
\newblock {\em CIT. Journal of Computing and Information Technology},
  17(1):67--80, 2009.

\bibitem{kragic2006strategies}
Danica Kragic and Marten Bjorkman.
\newblock Strategies for object manipulation using foveal and peripheral
  vision.
\newblock In {\em Computer Vision Systems, 2006 ICVS'06. IEEE International
  Conference on}, pages 50--50. IEEE, 2006.

\bibitem{krizhevsky2012imagenet}
Alex Krizhevsky, Ilya Sutskever, and Geoffrey~E Hinton.
\newblock Imagenet classification with deep convolutional neural networks.
\newblock In {\em Advances in neural information processing systems}, pages
  1097--1105, 2012.

\bibitem{lai2011large}
Kevin Lai, Liefeng Bo, Xiaofeng Ren, and Dieter Fox.
\newblock A large-scale hierarchical multi-view rgb-d object dataset.
\newblock In {\em Robotics and Automation (ICRA), 2011 IEEE International
  Conference on}, pages 1817--1824. IEEE, 2011.

\bibitem{lakshminarayanan2014mondrian}
Balaji Lakshminarayanan, Daniel~M Roy, and Yee~Whye Teh.
\newblock Mondrian forests: Efficient online random forests.
\newblock In {\em Advances in Neural Information Processing Systems}, pages
  3140--3148, 2014.

\bibitem{lauri2014stochastic}
Mikko Lauri and Risto Ritala.
\newblock Stochastic control for maximizing mutual information in active
  sensing.
\newblock In {\em IEEE Int. Conf. on Robotics and Automation (ICRA) Workshop on
  Robots in Homes and Industry}, 2014.

\bibitem{2016arXiv160301976L}
G.~{Li} and Y.~{Yu}.
\newblock {Deep Contrast Learning for Salient Object Detection}.
\newblock {\em ArXiv e-prints}, March 2016.

\bibitem{2016ITIP...25.5012L}
G.~{Li} and Y.~{Yu}.
\newblock {Visual Saliency Detection Based on Multiscale Deep CNN Features}.
\newblock {\em IEEE Transactions on Image Processing}, 25:5012--5024, November
  2016.

\bibitem{liu2016ssd}
Wei Liu, Dragomir Anguelov, Dumitru Erhan, Christian Szegedy, Scott Reed,
  Cheng-Yang Fu, and Alexander~C Berg.
\newblock Ssd: Single shot multibox detector.
\newblock In {\em European conference on computer vision}, pages 21--37.
  Springer, 2016.

\bibitem{lopes2012exploration}
Manuel Lopes, Tobias Lang, Marc Toussaint, and Pierre-Yves Oudeyer.
\newblock Exploration in model-based reinforcement learning by empirically
  estimating learning progress.
\newblock In {\em Advances in Neural Information Processing Systems}, pages
  206--214, 2012.

\bibitem{massios1998best}
Nikolaos~A Massios, Robert~B Fisher, et~al.
\newblock {\em A best next view selection algorithm incorporating a quality
  criterion}.
\newblock Department of Artificial Intelligence, University of Edinburgh, 1998.

\bibitem{minut2001reinforcement}
Silviu Minut and Sridhar Mahadevan.
\newblock A reinforcement learning model of selective visual attention.
\newblock In {\em Proceedings of the fifth international conference on
  Autonomous agents}, pages 457--464. ACM, 2001.

\bibitem{nguyen2013learning}
Sao~Mai Nguyen, Serena Ivaldi, Natalia Lyubova, Alain Droniou, Damien
  Gerardeaux-Viret, David Filliat, Vincent Padois, Olivier Sigaud, and
  Pierre-Yves Oudeyer.
\newblock Learning to recognize objects through curiosity-driven manipulation
  with the icub humanoid robot.
\newblock In {\em Development and Learning and Epigenetic Robotics (ICDL), 2013
  IEEE Third Joint International Conference on}, pages 1--8. IEEE, 2013.

\bibitem{oquab2015object}
Maxime Oquab, L{\'e}on Bottou, Ivan Laptev, and Josef Sivic.
\newblock Is object localization for free?-weakly-supervised learning with
  convolutional neural networks.
\newblock In {\em Proceedings of the IEEE Conference on Computer Vision and
  Pattern Recognition}, pages 685--694, 2015.

\bibitem{osswald2016speeding}
Stefan O{\ss}wald, Maren Bennewitz, Wolfram Burgard, and Cyrill Stachniss.
\newblock Speeding-up robot exploration by exploiting background information.
\newblock {\em IEEE Robotics and Automation Letters}, 1(2):716--723, 2016.

\bibitem{oudeyer2007intrinsic}
P-Y Oudeyer, Fr{\'e}d{\'e}ric Kaplan, and Verena~Vanessa Hafner.
\newblock Intrinsic motivation systems for autonomous mental development.
\newblock {\em Evolutionary Computation, IEEE Transactions on}, 11(2):265--286,
  2007.

\bibitem{Pan_2017_SalGAN}
Junting Pan, Cristian Canton, Kevin McGuinness, Noel~E. O'Connor, Jordi Torres,
  Elisa Sayrol, and Xavier~and Giro-i Nieto.
\newblock Salgan: Visual saliency prediction with generative adversarial
  networks.
\newblock In {\em arXiv}, January 2017.

\bibitem{papon2013voxel}
Jeremie Papon, Alexey Abramov, Markus Schoeler, and Florentin Worgotter.
\newblock Voxel cloud connectivity segmentation-supervoxels for point clouds.
\newblock In {\em Proceedings of the IEEE Conference on Computer Vision and
  Pattern Recognition}, pages 2027--2034, 2013.

\bibitem{peng2014rgbd}
Houwen Peng, Bing Li, Weihua Xiong, Weiming Hu, and Rongrong Ji.
\newblock Rgbd salient object detection: A benchmark and algorithms.
\newblock In {\em Computer Vision--ECCV 2014}, pages 92--109. Springer, 2014.

\bibitem{Pinheiro15}
P.~O. {Pinheiro}, R.~{Collobert}, and P.~{Dollar}.
\newblock {Learning to Segment Object Candidates}.
\newblock {\em ArXiv e-prints}, June 2015.

\bibitem{Pinheiro16}
P.~O. {Pinheiro}, T.-Y. {Lin}, R.~{Collobert}, and P.~{Doll{\`a}r}.
\newblock {Learning to Refine Object Segments}.
\newblock {\em ArXiv e-prints}, March 2016.

\bibitem{potapova2014attention}
Ekaterina Potapova, Karthik~M Varadarajan, Andreas Richtsfeld, Michael Zillich,
  and Markus Vincze.
\newblock Attention-driven object detection and segmentation of cluttered table
  scenes using 2.5 d symmetry.
\newblock In {\em 2014 IEEE International Conference on Robotics and Automation
  (ICRA)}, pages 4946--4952. IEEE, 2014.

\bibitem{rasolzadeh2010active}
Babak Rasolzadeh, M{\aa}rten Bj{\"o}rkman, Kai Huebner, and Danica Kragic.
\newblock An active vision system for detecting, fixating and manipulating
  objects in the real world.
\newblock {\em The International Journal of Robotics Research},
  29(2-3):133--154, 2010.

\bibitem{ren2015faster}
Shaoqing Ren, Kaiming He, Ross Girshick, and Jian Sun.
\newblock Faster r-cnn: Towards real-time object detection with region proposal
  networks.
\newblock In {\em Advances in Neural Information Processing Systems}, pages
  91--99, 2015.

\bibitem{saffari2009line}
Amir Saffari, Christian Leistner, Jakob Santner, Martin Godec, and Horst
  Bischof.
\newblock On-line random forests.
\newblock In {\em Computer Vision Workshops (ICCV Workshops), 2009 IEEE 12th
  International Conference on}, pages 1393--1400. IEEE, 2009.

\bibitem{Santos2016}
João~Machado Santos, Tomáš Krajník, and Tom Duckett.
\newblock Spatio-temporal exploration strategies for long-term autonomy of
  mobile robots.
\newblock {\em Robotics and Autonomous Systems}, 2016.

\bibitem{schmidhuber1991curious}
J{\"u}rgen Schmidhuber.
\newblock Curious model-building control systems.
\newblock In {\em Neural Networks, 1991. 1991 IEEE International Joint
  Conference on}, pages 1458--1463. IEEE, 1991.

\bibitem{schmidhuber2010formal}
J{\"u}rgen Schmidhuber.
\newblock Formal theory of creativity, fun, and intrinsic motivation
  (1990--2010).
\newblock {\em Autonomous Mental Development, IEEE Transactions on},
  2(3):230--247, 2010.

\bibitem{shi2016hierarchical}
Jianping Shi, Qiong Yan, Li~Xu, and Jiaya Jia.
\newblock Hierarchical image saliency detection on extended cssd.
\newblock {\em IEEE transactions on pattern analysis and machine intelligence},
  38(4):717--729, 2016.

\bibitem{van2012seeds}
Michael Van~den Bergh, Xavier Boix, Gemma Roig, Benjamin de~Capitani, and Luc
  Van~Gool.
\newblock Seeds: Superpixels extracted via energy-driven sampling.
\newblock In {\em Computer Vision--ECCV 2012}, pages 13--26. Springer, 2012.

\bibitem{vijayakumar2001overt}
Sethu Vijayakumar, J{\"o}rg Conradt, Tomohiro Shibata, and Stefan Schaal.
\newblock Overt visual attention for a humanoid robot.
\newblock In {\em Intelligent Robots and Systems, 2001. Proceedings. 2001
  IEEE/RSJ International Conference on}, volume~4, pages 2332--2337. IEEE,
  2001.

\bibitem{watkins1992q}
Christopher~JCH Watkins and Peter Dayan.
\newblock Q-learning.
\newblock {\em Machine learning}, 8(3-4):279--292, 1992.

\bibitem{weng2001autonomous}
Juyang Weng, James McClelland, Alex Pentland, Olaf Sporns, Ida Stockman,
  Mriganka Sur, and Esther Thelen.
\newblock Autonomous mental development by robots and animals.
\newblock {\em Science}, 291(5504):599--600, 2001.

\bibitem{zhang2013saliency}
Jianming Zhang and Stan Sclaroff.
\newblock Saliency detection: a boolean map approach.
\newblock In {\em Computer Vision (ICCV), 2013 IEEE International Conference
  on}, pages 153--160. IEEE, 2013.

\bibitem{zhao2011learning}
Qi~Zhao and Christof Koch.
\newblock Learning a saliency map using fixated locations in natural scenes.
\newblock {\em Journal of vision}, 11(3):9, 2011.

\bibitem{zhou2015cnnlocalization}
B.~Zhou, A.~Khosla, Lapedriza. A., A.~Oliva, and A.~Torralba.
\newblock {Learning Deep Features for Discriminative Localization.}
\newblock {\em CVPR}, 2016.

\bibitem{zhu2012unsupervised}
Jun-Yan Zhu, Jiajun Wu, Yichen Wei, Eric Chang, and Zhuowen Tu.
\newblock Unsupervised object class discovery via saliency-guided multiple
  class learning.
\newblock In {\em Computer Vision and Pattern Recognition (CVPR), 2012 IEEE
  Conference on}, pages 3218--3225. IEEE, 2012.

\bibitem{zitnick2014edge}
C~Lawrence Zitnick and Piotr Doll{\'a}r.
\newblock Edge boxes: Locating object proposals from edges.
\newblock In {\em Computer Vision--ECCV 2014}, pages 391--405. Springer, 2014.

\end{thebibliography}


\begin{IEEEbiography}[{\includegraphics[width=1in,height=1.25in,clip,keepaspectratio]{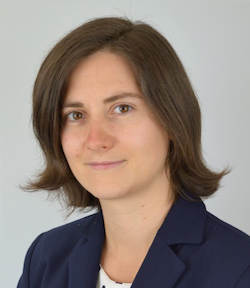}}]{Celine Craye}
C\'eline Craye received the diplome d'Ing\'enieur in telecommunications from the Ecole Nationale Sup\'erieure des T\'el\'ecommunications de Bretagne, France, in 2012. She received her M.A.Sc. degree from the Electrical and Computer Engineering Department at the University of Waterloo in 2013, and her Ph.D. from the University of Paris Saclay in 2017. She is now a research engineer at the Thales VisionLab, the Vision innovation laboratory for video-surveillance of the Thales Group. Her research interests are in computer vision, machine learning and developmental robotics.
\end{IEEEbiography}
\begin{IEEEbiography}[{\includegraphics[width=1in,height=1.25in,clip,keepaspectratio]{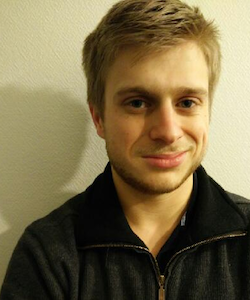}}]{Timoth\'ee Lesort}
Timoth\'ee Lesort received the diplome d'Ing\'enieur in electronics with robotics and learning algorithms major from the Ecole CPE Lyon, France, in 2017.
He is currently a Ph.D. candidate at ENSTA Paristech in France at the UIIS laboratory.  The Ph.D. is granted by the CIFRE program in partnership with Thales Vision Lab. His research interests are in computer vision, incremental deep learning and developmental robotics.
\end{IEEEbiography}

\begin{IEEEbiography}[{\includegraphics[width=1in,height=1.25in,clip,keepaspectratio]{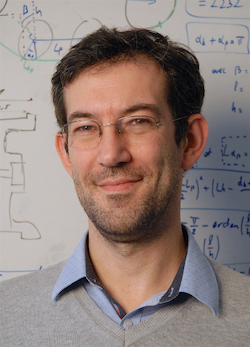}}]{David Filliat}
David Filliat graduated from the Ecole Polytechnique in 1997 and obtained a PhD on bio-inspired robotics navigation from Paris VI university in 2001. After 4 years as an expert for the robotic programs in the French armament procurement agency, he is now professor at Ecole Nationale Sup\'erieure de Techniques Avanc\'ees ParisTech. Head of the Robotics and Computer Vision team since 2006, he obtained the Habilitation \'a Diriger des Recherches en 2011. He is also a member of the ENSTA ParisTech INRIA FLOWERS team. His main research interest are perception, navigation and learning in the frame of the developmental approach for autonomous mobile robotics. \url{http://www.ensta-paristech.fr/~filliat/}
\end{IEEEbiography}

%\newpage

\begin{IEEEbiography}[{\includegraphics[width=1in,height=1.25in,clip,keepaspectratio]{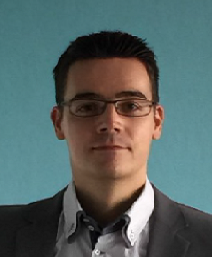}}]{Jean-Fran\'cois Goudou}
Jean-Fran\'cois Goudou was born in Versailles, France, in 1979. He received the M.E. degree in applied mathematics from the Ecole Polytechnique, Palaiseau, France, in 2004, and the Ph.D. degrees in computer vision from Telecom ParisTech, Paris, France, in 2007.
In 2008, he joined the Advanced studies department Theresis, from Thales, as a project manager for collaborative national and European projects. He is since 2009 in charge of the demonstrators of the VisionLab, the Vision innovation laboratory for video-surveillance of the Thales Group. He is also leading several research projects in the topics of video-surveillance and algorithmic evaluation. He is since 2016 deputy head of the VisionLab, in charge of academic co-operations and H2020 projects proposals and management. His research interest include the complete sensing chain from camera to processing, the human vision and bio-inspired processing and neural networks, including deep learning.
\end{IEEEbiography}


%\vfill


\end{document}